\documentclass[10pt,twocolumn,letterpaper]{article}

\usepackage[pagenumbers]{arxiv}

\usepackage{graphicx}
\usepackage{afterpage}
\usepackage{amsmath}
\usepackage{amssymb}
\usepackage{booktabs}
\usepackage{gensymb}
\usepackage{hyphenat}
\usepackage{fancybox}
\usepackage[breaklinks,colorlinks]{hyperref}
\usepackage{multirow}

\usepackage[capitalize]{cleveref}

\begin{document}

\title{Fast Facial Landmark Detection and Applications: A Survey}

\author{Kostiantyn Khabarlak \\
Dnipro University of Technology\\
Ukraine\\
{\tt\small habarlack@gmail.com} \\
\and Larysa Koriashkina\\
Dnipro University of Technology\\
Ukraine
}

\maketitle

\begin{abstract} Dense facial landmark detection is one of the key elements of face processing pipeline. It is used in virtual face reenactment, emotion recognition, driver status tracking, etc. Early approaches were suitable for facial landmark detection in controlled environments only, which is clearly insufficient. Neural networks have shown an astonishing qualitative improvement for in-the-wild face landmark detection problem, and are now being studied by many researchers in the field. Numerous bright ideas are proposed, often complimentary to each other. However, exploration of the whole volume of novel approaches is quite challenging. Therefore, we present this survey, where we summarize state-of-the-art algorithms into categories, provide a comparison of recently introduced in-the-wild datasets (e.g., 300W, AFLW, COFW, WFLW) that contain images with large pose, face occlusion, taken in unconstrained conditions. In addition to quality, applications require fast inference, and preferably on mobile devices. Hence, we include information about algorithm inference speed both on desktop and mobile hardware, which is rarely studied. Importantly, we highlight problems of algorithms, their applications, vulnerabilities, and briefly touch on established methods. We hope that the reader will find many novel ideas, will see how the algorithms are used in applications, which will enable further research.
\end{abstract}


\blfootnote{
  This is the peer reviewed version of the following article: \nohyphens{K.~Khabarlak, L.~Koriashkina \emph{Fast Facial Landmark Detection and Applications: A Survey}. Journal of Computer Science \& Technology, vol. 22, no. 1, pp. 12–41, 2022.}, which has been published in final form at \href{https://doi.org/10.24215/16666038.22.e02}{DOI:~10.24215/16666038.22.e02}. This article may be used for non-commercial purposes in accordance with JCS\&T Terms and Conditions.
}

\section{Introduction}

Dense facial landmark detection is one of the key elements of face processing pipeline. Applications include virtual face animation, emotion recognition, driver status tracking, etc. Early attempts to solve the problem were based on deformable face model, where statistical algorithms predicted face model deformation coefficients. These approaches were unsuitable for landmark annotation with large pose, face occlusion or unusual illumination. Later, attention has been driven to neural networks, that show high quality in solving tasks, in which we, humans, are good at, such as image classification or natural language processing. Neural networks have also shown an astonishing qualitative improvement for in-the-wild face landmark detection problem, and are now being actively studied by many researchers in the field. Primarily, neural networks were designed to be executed on servers with many GPUs and a stable power supply. However, the development of Internet of Things and mobile devices makes client-server applications sometimes impractical or even unacceptable. For example, when Internet connectivity is poor, low latency data processing is required, if the amounts of raw data generated are too large to be sent over to a server. Finally, when no data can leave the user's device for security reasons. In many of these cases use of neural networks is desirable, and processing should be done directly on a mobile device. Thus, on-device machine learning has become one of the most prominent machine learning research directions~\cite{ShuffleNet,MobileNetV2}.

In this paper we present a description of recently introduced neural-network-based facial landmark detection algorithms. Existing surveys are quite old and mostly cover either statistical algorithms or the ones based on ensembles of regression trees~\cite{FacialFeaturePointDetectionSurvey2018,FacialLandmarkDetectionSurvey2019}. These algorithms show poor facial landmark detection quality for in-the-wild pictures (i.e., taken in unconstrained environments). Recently, numerous bright neural-network-based approaches were proposed, that show substantially better quality. However, exploration of the whole volume of novel approaches is quite challenging. Therefore, we present this work. The primary focus of this survey is on recently introduced algorithms, covering years 2018 -- 2021. We include some important older algorithm for completeness as well.

We start our survey by defining facial landmark detection problem, algorithm quality assessment metrics. Next, we describe common in-the-wild datasets (e.g., 300W, AFLW, COFW, WFLW) with dense landmark annotation (from 21 to 98 landmarks). These datasets contain images taken in unconstrained conditions with large pose, face occlusion, different emotions, etc. The following section describes ideas of facial landmark algorithms, that have led to accuracy improvement or have proposed a novel way to solve the problem. This section is key for this survey. To make algorithm ideas clear, we start by explaining common neural network backbones used for facial landmark detection. Based on these materials, we follow with an explanation of facial landmark detection algorithms ordered by years. Finally, we summarize state-of-the-art algorithms into categories, provide accuracy comparison on recently introduced in-the-wild datasets. In addition to quality, applications require fast inference, possibly on mobile devices. Hence, we include information about algorithm inference speed both on desktop and mobile hardware, which is rarely studied in literature. Where available, inference time is shown for desktop CPU and GPU, as well as mobile phone. Also, we provide estimated number of neural network parameters and floating-point operations. These are the metrics, that influence memory consumption and inference time correspondingly. Importantly, we highlight problems of algorithms, their applications and vulnerabilities. Overall, we note that algorithm accuracy needs to be improved by the next generation of algorithms. Also, state-of-the-art algorithms have inference times that are quite high for practical applications. We hope that in this survey the reader will find many novel ideas, will see how the algorithms are used in applications, which will enable new research in the field.

The paper is structured as follows: \cref{sec:problem-statement} covers facial landmark detection problem. \cref{sec:datasets} describes datasets used to train and evaluate models. \cref{sec:early-landmark-detection} gives a brief introduction of historical landmark detection methods. The main \cref{sec:backbones,sec:main-landmark-detection} cover common neural network backbones and landmark detection algorithms correspondingly. Analysis of algorithm accuracy, inference speed, and summary of novel ideas is presented in \cref{sec:algorithm-summary}. \cref{sec:landmark-detection-applications} is focused on real-world use of face landmark detection methods. We show several possible approaches of joint face and landmark detection algorithms in \cref{sec:joint-face-landmark-detection}.  Applications of dense facial landmark detection are shown in animation in \cref{sec:face-animation}, driver status tracking in \cref{sec:driver-tracking}, face and emotion recognition in \cref{sec:face-recognition}. Finally, adversarial attack vulnerability is discussed in \cref{sec:vulnerabilities}.

\section{Facial Landmark Detection Problem Statement}\label{sec:problem-statement}

\begin{figure}[htb]
  \captionsetup[subfigure]{justification=centering}
  \centering
  \begin{subfigure}{0.32\linewidth}
    \centering
    \includegraphics[height=0.9\linewidth]{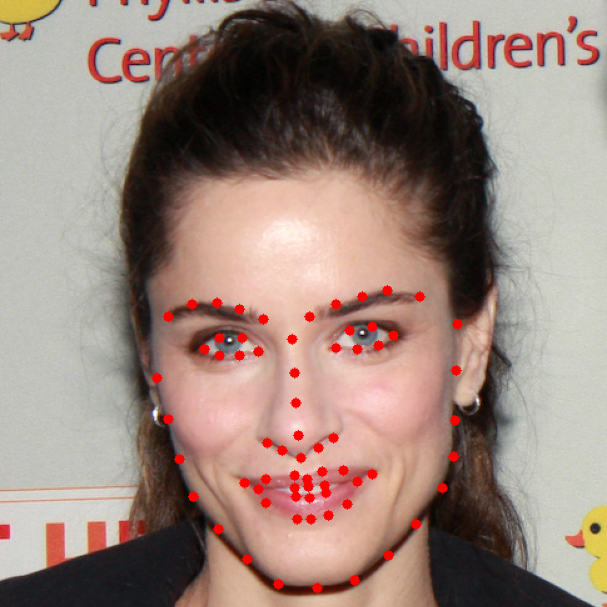}
    \caption{300W\protect\footnotemark}\label{fig:datasets-300w}
  \end{subfigure}
  \begin{subfigure}{0.32\linewidth}
    \centering
    \includegraphics[width=0.9\linewidth]{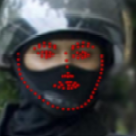}
    \caption{WFLW\protect\footnotemark~\cite{LAB}}\label{fig:datasets-wflw}
  \end{subfigure}
  \caption{Examples of faces annotated with facial landmarks from several of the commonly used datasets: 300W and AFLW. In both cases landmarks cover areas of jaw, eyes, eyebrows, nose, lips. However, annotation schemes differ. For instance, WFLW annotates lower and upper boundary of eyebrows, whereas 300W has a single central line. Also, WFLW has the densest landmark annotation.}\label{fig:datasets}
\end{figure}

Let \(I\) be an input image, which is represented in a form of 3-dimensional tensor of size \(W \times H \times C\), where \(W\), \(H\), \(C\) are the width, height, and number of image color channels correspondingly. Note, that typically color images are used with 3 channels, one for red, green and blue colors. Then facial landmark detection problem is to find such function \(\Phi: I \rightarrow Y\), that from the input image \(I\) predicts a landmark matrix \(\hat{Y} \in R^{N_L \times 2}\), where \(N_L\) is the number of facial landmarks, \(\hat{Y}_{i\,1} \in [0; W]\) represents \(X\) coordinate and \(\hat{Y}_{i\,2} \in [0; H]\) represents \(Y\) coordinate of \(i\)\textsuperscript{th} landmark. Number of facial landmarks \(N_L\) and exact mapping between \(i\)\textsuperscript{th} facial landmark and its location on the face (the so-called annotation scheme) are defined at dataset level. Examples of face landmark annotations are present in \cref{fig:datasets}. Also, dataset defines which images are used to train function \(\Phi\) (train set) and which to evaluate (test set).

\footnotetext[1]{Image is based on \href{https://commons.wikimedia.org/wiki/File:Amanda_Peet_at_Light_Up_a_Life.jpg\#/media/File:Amanda_Peet_at_Light_Up_a_Life.jpg}{this source}. License: CC BY 2.0. It has been annotated with landmarks available in 300W dataset by the authors of this survey.}
\footnotetext{Author's written consent has been acquired to include this image.}

Next, we present commonly used metrics to report algorithms' quality on facial landmark detection datasets. Note, that each dataset has a special protocol, which defines train/test split, metrics for algorithm comparison, etc. The main metrics include~\cite{LUVLi,AWing}:

\begin{enumerate}
  \item Normalized Mean Error (NME, \%):
        \begin{align}
          \label{eq:NME}
          NME & = \frac{1}{K} \sum_{k=1}^{K} NME_k, \notag
          \\NME_k &= \frac{1}{N_L} \sum_{i=1}^{N_L} \frac{||Y_i - \hat{Y}_i||}{d} \times 100,
        \end{align}
        where \(Y\) is the matrix of true landmark locations, \(\hat{Y}\) is the matrix of predicted landmark locations, \(d\) is the normalization coefficient (different for each dataset), \(N_L\) is the number of facial landmarks per face in the dataset, \(K\) is the number of images in the test set. Lower metric values are better.

  \item Failure Rate (FR, \%):

        \begin{equation}
          \label{eq:FailureRate}
          FR = \frac{1}{K} \sum_{k=1}^{K}{[NME_k \ge 10\%]} \times 100,
        \end{equation}

        denotes number of images with Normalized Mean Error above 10 \% threshold. Lower metric values are better.

  \item Cumulative Error Distribution -- Area Under Curve (CED-AUC). First, fraction of images whose NME is less than or equal to the NME value on X axis is plotted. Area under curve is then computed. Typically, NME is taken in range \([0; 10 \%]\). Computed CED-AUC value is always scaled in range \([0; 1]\). Greater metric values are better, and denote that larger part of the test set is well predicted.
\end{enumerate}

\section{Common Face Landmark Datasets}\label{sec:datasets}

There are several open datasets available to train and evaluate quality of face landmark detection algorithms. Each of the datasets includes image of a person and corresponding face landmark annotations. Landmarks are provided in a separate file. The datasets can include photos of the following kinds:
\begin{itemize}
  \item in controlled environment (e.g., studio) or in-the-wild;
  \item with different shooting conditions, such as presence of face occlusion, large pose, make-up, etc.;
  \item real images or synthetic (when faces are generated with an algorithm);
  \item 2D or 3D face landmarks.
\end{itemize}

Next, we describe typical datasets used to train and evaluate facial landmark detection models. The datasets were selected from the following sources: 1)~introduced jointly with a novel facial landmark detection algorithm; 2)~presented separately, but at least one of the algorithms from \cref{sec:main-landmark-detection} uses the dataset for training or evaluation. While we discuss all such datasets, the focus of this survey is on in-the-wild 2D face landmark datasets with non-synthetic images. Note, that in-the-wild datasets also include images in controlled environments, which makes them applicable to a wide range of practical use-cases.

\textbf{300 Faces in-the-Wild (300W)}~\cite{300WDatasetChallenge} dataset contains a collection of different datasets: LFPW~\cite{LFPW}, AFW~\cite{AFW}, HELEN~\cite{HELEN}, XM2VTS~\cite{XM2VTS} and IBUG, that were relabeled with 68 facial landmarks. The protocol defines which images should be used for training and which for testing. The testing subset is split into \textit{common}, \textit{challenge} and \textit{full}. Normalized Mean Error for each of the splits is usually presented for comparison. The NME is normalized ($d$ in \cref{eq:NME}) by inter-pupil or inter-ocular (outer eye corner) distance. This is done, so that faces of different sizes make an equal contribution to the resulting error. Note, that images in the 300W dataset have different shooting conditions (lighting, color gamut), emotions and faces at an angle. There have been multiple extensions to the 300W datasets presented: \textbf{300W-LP-2D}~\cite{FaceAlignmentLargePose3D}, where the original 300W dataset has been expanded with synthetically generated images with large pose; \textbf{Masked-300W}~\cite{SAAT} has synthetically added medical mask to the 300W dataset images. However, the same blue medical mask model has been used for all images, which is a disadvantage.

\textbf{Annotated Facial Landmarks in-the-Wild (AFLW)}~\cite{AFLWDataset} contains a larger number of images, yet they are labeled with only 21 facial landmarks. In comparison to 300W, this dataset has face photographs taken at a larger angle in range of $\pm 120 \degree$ yaw and $\pm 90 \degree$ pitch. The authors propose splitting the dataset into AFLW-Frontal (with face photos that are close to frontal) and AFLW-Full (all images). Also, there is a relabeled version with $68$ facial landmarks, named \textbf{AFLW-68}~\cite{AVS-AFLW-68Dataset}, yet in practice it is used less often. \textbf{MERL-RAV} dataset presented in~\cite{LUVLi} has AFLW relabeled with 68 landmarks with an extra visibility label, such as: 1) visible; 2) self-occluded (for instance, due to large pose); 3) occluded by other object (hand, etc.). NME metric, normalized by face bounding box size (diagonal), is used for comparisons.

\textbf{Caltech Occluded Faces in-the-Wild (COFW)}~\cite{COFW} focuses on face images, that are partially occluded by real-world objects (microphone, etc.) or by the person itself (hair, hand, etc.). In addition to the NME metric, Failure Rate (FR, \cref{eq:FailureRate}) is used. The dataset has 29 landmark annotations. NME is normalized by either inter-pupil or inter-ocular distance. The COFW test set has also been relabeled to 68 landmarks in \textbf{COFW-68}~\cite{COFW-68}, but no COFW training set with 68 landmarks is available. COFW-68 can be used to assess landmark detection quality, when the network has been trained on a different dataset.

\textbf{Wider Facial Landmarks in-the-Wild (WFLW)}~\cite{LAB} is the dataset with the largest number of facial landmarks (98 landmarks). It is also the most recently introduced. In comparison to the datasets covered so far, WFLW has more images taken under unusual conditions, e.g., with make-up, wide range of emotions, poses, in various lighting conditions, etc. All three above-mentioned metrics are used to present the results: NME, Failure Rate and CED-AUC.  NME is normalized by inter-ocular distance. The results are reported for each subset of unusual images, as well as for all images available in the dataset. This makes it possible to analyse, which conditions are the most challenging to the algorithms. The following subsets are available in WFLW dataset: Pose, Expression, Illumination, Make-Up, Occlusion, Blur. Information about image scene type is included in the dataset and can also be used during training.

\textbf{Menpo-2D}~\cite{Menpo2D,Menpo2D3D} is a collection of frontal and profile faces. However, annotation schemes and number of landmarks are different between types of faces. The dataset is less used in practice.

In addition, there are many datasets that provide 3D annotations of facial landmarks (either synthetically generated or manually), such as 300W-LP~\cite{FaceAlignmentLargePose3D}, AFLW2000-3D~\cite{FaceAlignmentLargePose3D}, LS3D-W~\cite{FAN}, Menpo-3D~\cite{Menpo3D,Menpo2D3D}. Also, landmark annotated video datasets exist, e.g., 300 Videos in the Wild (300VW)~\cite{300VW1,300VW2,300VW3}.

\cref{tab:facial-landmark-detection-datasets} summarizes information about common datasets. We include information about number of labeled images for algorithm training and testing, as well as number of facial landmarks the dataset has been labeled with. The most widely used datasets are shown in bold.

\begin{table}[htb]
  \caption{Information about facial landmark datasets: number of images contained in training and testing tests, as well as number of facial landmarks the dataset has been annotated with.}\label{tab:facial-landmark-detection-datasets}
  \centering
  \begin{tabular}{lrrr}
    \toprule
    Dataset                                    & Train  & Test  & Land. \\
    \toprule
    \textbf{300W}~\cite{300WDatasetChallenge}  & 3,837  & 600   & 68    \\
    \textbf{AFLW}~\cite{AFLWDataset}           & 20,000 & 4,386 & 21    \\
    \textbf{COFW}~\cite{COFW}                  & 1,345  & 507   & 29    \\
    \textbf{WFLW}~\cite{LAB}                   & 7,500  & 2,500 & 98    \\
    300W-LP-2D~\cite{FaceAlignmentLargePose3D} & 61,225 & -     & 68    \\
    AFLW-68~\cite{AVS-AFLW-68Dataset}          & 20,000 & 4,386 & 68    \\
    COFW-68~\cite{COFW-68}                     & -      & 507   & 68    \\
    Menpo-2D~\cite{Menpo2D,Menpo2D3D}          & 7,564  & 7,281 & 68/39 \\
    MERL-RAV~\cite{LUVLi}                      & 15,449 & 3,865 & 68    \\
    Masked-300W~\cite{SAAT}                    & 3,837  & 600   & 68    \\
  \end{tabular}
\end{table}

\section{Facial Landmark Detection Algorithms}

\subsection{Early Landmark Detection Algorithms}\label{sec:early-landmark-detection}

First algorithms were mainly based on fitting a deformable face mesh. The most prominent algorithms include Active Shape Model (ASM), Active Appearance Model (AAM) and Constrained Local Model (CLM)~\cite{FacialFeaturePointDetectionSurvey2018, FacialLandmarkDetectionSurvey2019}. Based on the obtained mesh, each of the landmark locations are computed. In many cases such algorithms utilize statistical methods as a base. They have good enough prediction accuracy in controlled environments (with proper lighting and frontal face). However, such approaches offer underwhelming performance for most types of in-the-wild images. Next wave of methods was based on Random Forests and Gradient Boosting, such as ERT~\cite{ERT} algorithm, which we describe below. Such methods have better accuracy, but performance for occluded faces, faces shot under large angle or unusual illumination is still insufficient. As will be shown later in this work, many practical applications require accurate in-the-wild facial landmark detection.

\textbf{Dlib}~\cite{Dlib} is an open-source machine learning library. Among others, it has \textbf{Ensemble of Regression Trees (ERT)}~\cite{ERT} facial landmark detection algorithm, which is a cascade, based on gradient boosting. The authors use a ``mean'' face template as an initial approximation, then the template is refined over several iterations. The algorithm requires the face to be first detected in the frame (Viola-Jones~\cite{ViolaJones} face detector is used). Note, that most facial landmark detection algorithms require face to be first detected. High speed is the main advantage of ERT (according to the authors, around 1 millisecond per face). The library contains ERT implementation, trained on 300W dataset. The algorithm is still actively used in the modern research thanks to an open implementation and speed. However, not so long ago it has been shown that neural networks are preferred in terms of quality for faces with large pose~\cite{WingLoss}. Mobile-friendly implementations of ERT are available.

An overview of early neural-network-based facial landmark detection algorithms can be found in~\cite{FacialLandmarkDetectionSurvey2019, FrenchFacialLandmarkDetectionSurvey}.

\subsection{Face Landmark Detection Network Backbones}\label{sec:backbones}

\begin{figure*}[tp]
  \captionsetup[subfigure]{justification=centering}
  \centering
  \includegraphics[width=\linewidth]{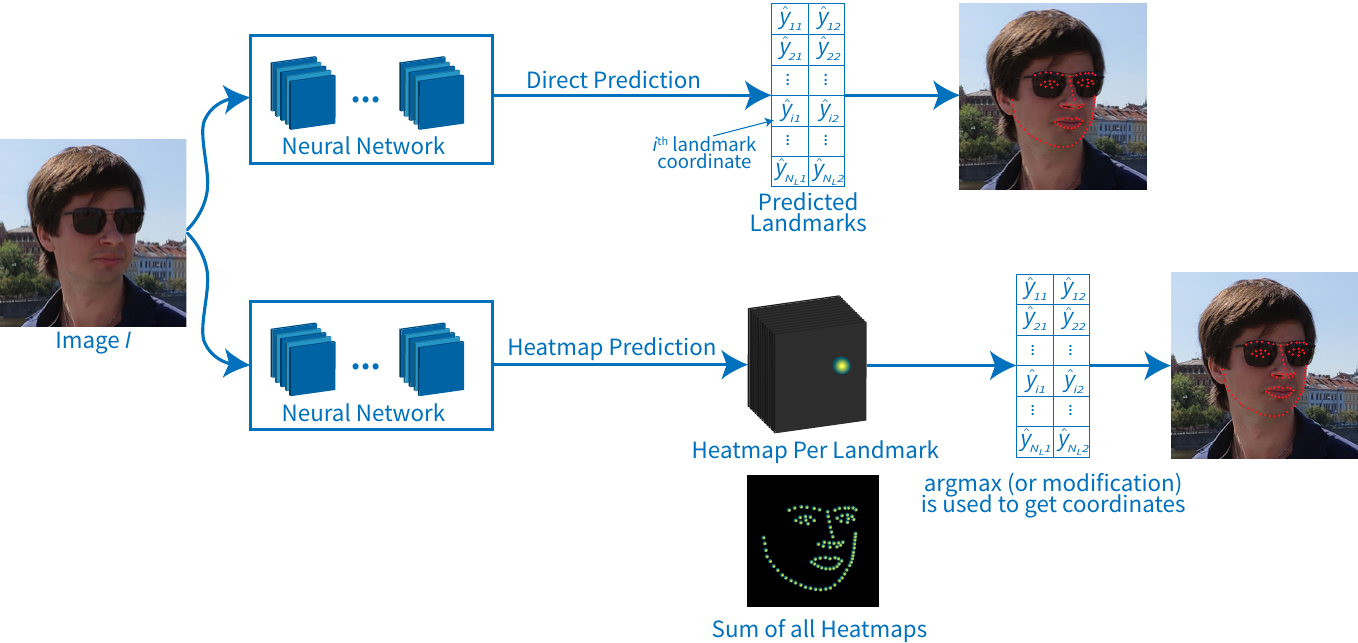}
  \caption{Direct landmark regression (upper row). The problem is solved in a form of regression, where actual landmark coordinates \((x, y)\) are predicted directly by the algorithm. Heatmap-based (bottom row). The algorithm predicts probability distributions of landmark locations in a form of heatmaps. One heatmap per each of the landmarks is formed. Argmax (or its modification) is used to get each landmark coordinates.}\label{fig:direct-vs-heatmap-prediction}
\end{figure*}

Modern in-the-wild face landmark detection algorithms are based on neural networks. They are divided into 2 main categories: \textit{direct} (or \textit{coordinate}) regression methods, when the model predicts \textit{x, y} coordinates directly for each landmark; \textit{heatmap}-based regression methods, where a 2D heatmap is built for each landmark. The values in the heatmap can be interpreted as probabilities of landmark location at a certain image location. Typically, argmax or its modification is used to acquire exact landmark coordinates from the heatmap. \cref{fig:direct-vs-heatmap-prediction} illustrates the two approaches. As neural network architectures have become more complex, algorithms typically base on a pre-defined network architecture (called backbone). Facial landmark detection algorithms, described in the following subsection, propose modifications to training, inference procedure or the backbone itself. Here we introduce main backbones for landmark detection problem. Note, that in many cases backbones for face landmark and human pose (whole body) landmark detection are the same. Direct regression methods typically use widely known backbones from ImageNet challenge~\cite{Imagenet}, such as ResNet~\cite{ResNet}, MobileNetV2~\cite{MobileNetV2}, MobileNetV3~\cite{MobileNetV3}, ShuffleNet-V2~\cite{ShuffleNetV2}. Heatmap-based methods commonly use Hourglass~\cite{Hourglass} network architecture, but also HRNet~\cite{HRNet} and CU-Net~\cite{CU-Net}. Such backbones are less known. Thus, we describe them here.

\begin{figure*}[tp]
  \captionsetup[subfigure]{justification=centering}
  \centering
  \includegraphics[width=\linewidth]{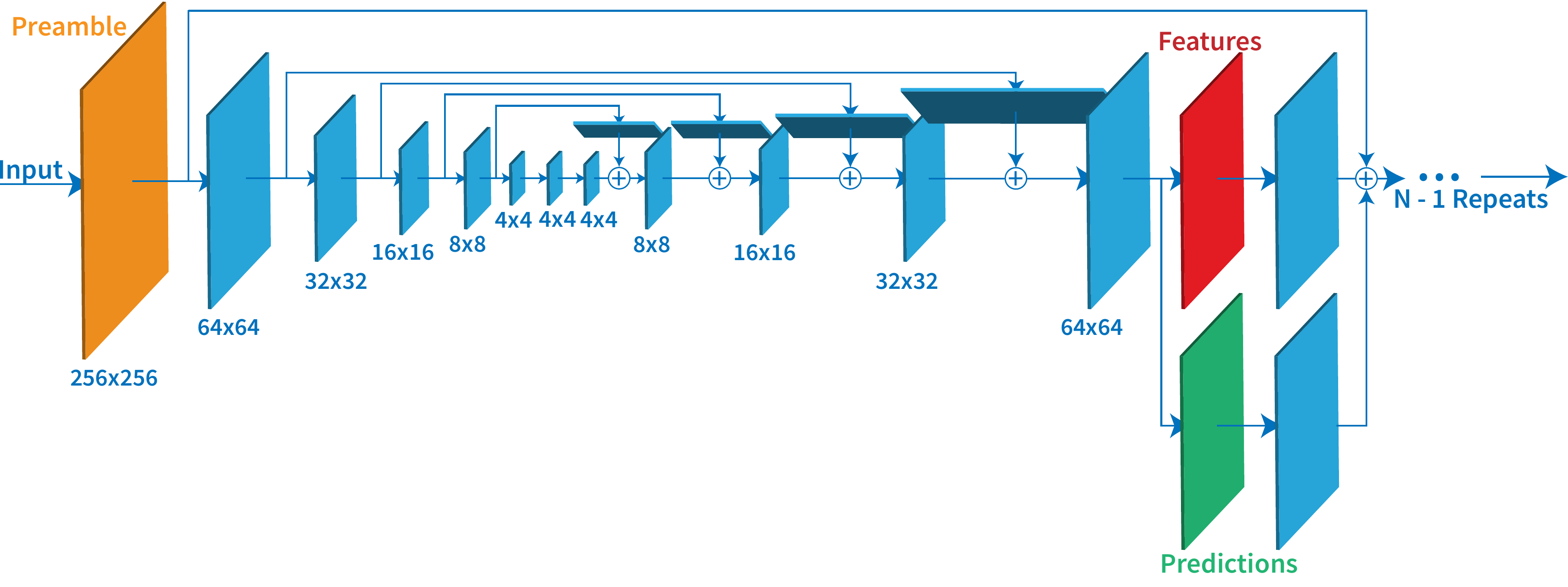}
  \caption{Hourglass architecture. Input image of size \(256 \times 256\) is rescaled to \(64 \times 64\) in Preamble (orange). Hourglass encoder-decoder module is to follow. First, Hourglass processes input via Residual blocks (blue boxes) and downscales features after each block (encoder part). Then in decoder features are upscaled after each Residual block. Additional skip connections between blocks of the same resolution are added to improve gradient propagation. Note, that encoder block's output is processed via an extra Residual block before signal is added to decoder features. Module features (in red) and predictions (in green) are formed after each Hourglass block. Then Hourglass block is repeated \(N - 1\) times for a stack of \(N\) Hourglass modules. At training time not only final, but also intermediate predictions participate in loss function computation (the so-called intermediate supervision).}\label{fig:hourglass}
\end{figure*}

\textbf{Hourglass}~\cite{Hourglass} architecture has been initially designed for human pose estimation. The network takes a \(256 \times 256\) image as an input. The authors note that processing the image at full resolution would require too much computation and memory. This is why a convolutional block is used to quickly process the image to obtain feature map of resolution \(64 \times 64\), which remains the maximum feature map resolution till the end of the network. The feature map is then processed via Hourglass modules. An illustration of Hourglass network is shown in \cref{fig:hourglass}. Note, that architecture allows \textit{stacking}, i.e., Hourglass modules can be repeated sequentially multiple times. Typically stacks of 1, 2 or 4 Hourglass modules are used. The network outputs heatmaps at a resolution of \(64 \times 64\), a single heatmap is produced for each of the landmarks.

Hourglass module follows encoder-decoder architecture. Input image is processed via convolutional blocks at different feature map resolutions. First, feature map resolution is decreased after each convolutional block (encoder part), then feature map resolution is restored (increased) after each block (decoder part). Accuracy of human pose estimation, facial landmark detection and several other tasks is improved by processing image at multiple resolutions.

Overall, stacked Hourglass architecture becomes quite deep, which slows down training. The authors propose two ideas to solve the problem: skip connections inside Hourglass module and intermediate supervision. Firstly, as is clearly seen from \cref{fig:hourglass}, after each convolutional block the output is split into two parts, one part is downscaled and fed into next convolutional block, another is \textit{skipped} from encoder to decoder. The latter concept is then referred to as ``skip connection''. This improves gradient propagation. Secondly, intermediate supervision is applied to each Hourglass module (as previously said, stacks of multiple Hourglass modules can be constructed). Prediction heatmaps are always constructed after each Hourglass module (shown in green in \cref{fig:hourglass}). The training loss includes weighted sum of losses for each of the heatmap predictions.

\textbf{CU-Net}~\cite{CU-Net} tries to improve Hourglass architecture not only in quality, but also in memory footprint and inference time. The authors note an importance of efficient architecture for use on mobile devices. Similarly to Hourglass, the network takes \(256 \times 256\) image as an input and resizes it in preamble to \(64 \times 64\), which remains the maximum resolution at which features are processed till the end of the network. To improve training and enable deeper CU-Net stacks, the authors propose to add skip connections not only between features of the same module, but also between different modules. To avoid excessive number of skip connections, a concept of Order-\(K\) coupling has been introduced in the paper. Order-\(K\) coupling denotes that skip connections will be added only \(K\) modules forward. In most cases, adding skip connections to one module forward (\(K=1\)) seems sufficient. The authors decrease memory consumption and improve inference speed by avoiding unnecessary features copies, sharing memory, and quantizing both features and parameters. In addition, blocks with smaller number of features are used to decrease overall number of parameters. All these improvements have allowed to achieve similar to Hourglass accuracy on human pose estimation with only a fraction of parameters and higher inference speed. Exact number of parameters and inference times are shown in \cref{tab:hourglass-vs-cu-net}. However, despite the improvements, most recently introduced approaches prefer Hourglass architecture over CU-Net as will be shown later.

\begin{table}[htb]
  \caption{Comparison of number of parameters and inference time for Hourglass and CU-Net architectures.~\cite{CU-Net}. Larger stacks still have reasonable inference time and small number of parameters, which is achieved thanks to memory sharing, quantization and smaller blocks.}\label{tab:hourglass-vs-cu-net}
  \centering
  \begin{tabular}{lrrr}
    \toprule
    Method               & \# Params (M) & Inference (ms) \\
    \toprule
    4\(\times\)Hourglass & 25.5          & 48.9           \\
    8\(\times\)CU-Net    & 7.9           & 36.1           \\
    16\(\times\)CU-Net   & 15.9          & 70.8
  \end{tabular}
\end{table}

\begin{figure*}[tp]
  \captionsetup[subfigure]{justification=centering}
  \centering
  \includegraphics[width=0.9\linewidth]{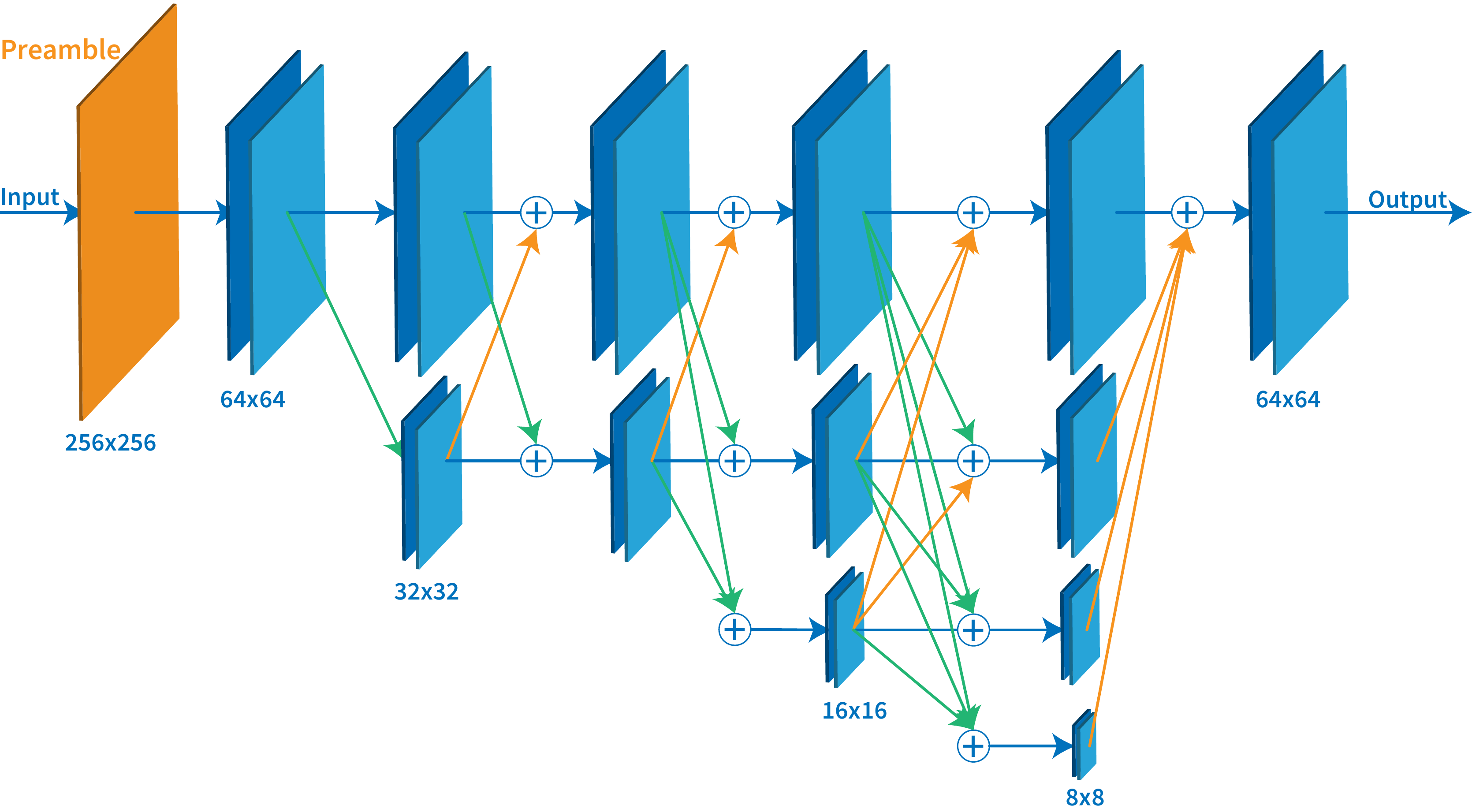}
  \caption{HRNet architecture. Input image of size \(256 \times 256\) is rescaled to \(64 \times 64\) in Preamble (orange). Next, input is processed at resolution of \(64 \times 64\). After each set of convolutional blocks, a parallel branch is added with 4 times smaller resolution. Overall, 4 parallel branches are created till the network end, with the smallest resolution of \(8 \times 8\). Features between blocks of different resolutions are exchanged. To pass features to blocks of higher resolution, nearest neighbor upsampling is used (orange arrows). To pass features to blocks of smaller resolution, strided convolution is used (green arrows). Blue arrows denote that feature map is not rescaled. The final output contains a sum of features of all scales.}\label{fig:hrnet}
\end{figure*}

\textbf{HRNet}~\cite{HRNet} has also been initially proposed for human pose estimation and then adapted to face landmark prediction in~\cite{HRNetV2} and other works. This architecture significantly differs from the previous two. HRNet doesn't follow encoder-decoder architecture and doesn't use multiple stacks. Instead, parallel branches with different feature resolutions are maintained throughout the network. An illustration of HRNet architecture is shown in \cref{fig:hrnet}. Similarly to previous works, the network takes an image of size \(256 \times 256\), which is then resized to \(64 \times 64\) feature map in preamble. Next, the image is processed via convolutional blocks. Then another branch of resolution \(32 \times 32\) is added. Note, that in contrast to previous works, \(64 \times 64\) branch continues to be processed in parallel. The authors propose to exchange features between parallel branches. However, these feature maps are of different resolutions. To downscale feature map, strided convolution is used. To upscale feature map, nearest neighbor upsampling is used. Till the end of the network 4 parallel branches with different feature map resolutions are created. The final heatmap is generated at resolution of \(64 \times 64\). At a similar number of parameters to a stack of 8 Hourglass modules, HRNet uses nearly twice fewer floating-point operations. Additionally, HRNet network width (number of convolutional channels) can be configured to change overall number of parameters and resulting inference speed.

\begin{figure*}[tp]
  \captionsetup[subfigure]{justification=centering}
  \centering
  \begin{subfigure}{0.48\linewidth}
    \includegraphics[width=\linewidth]{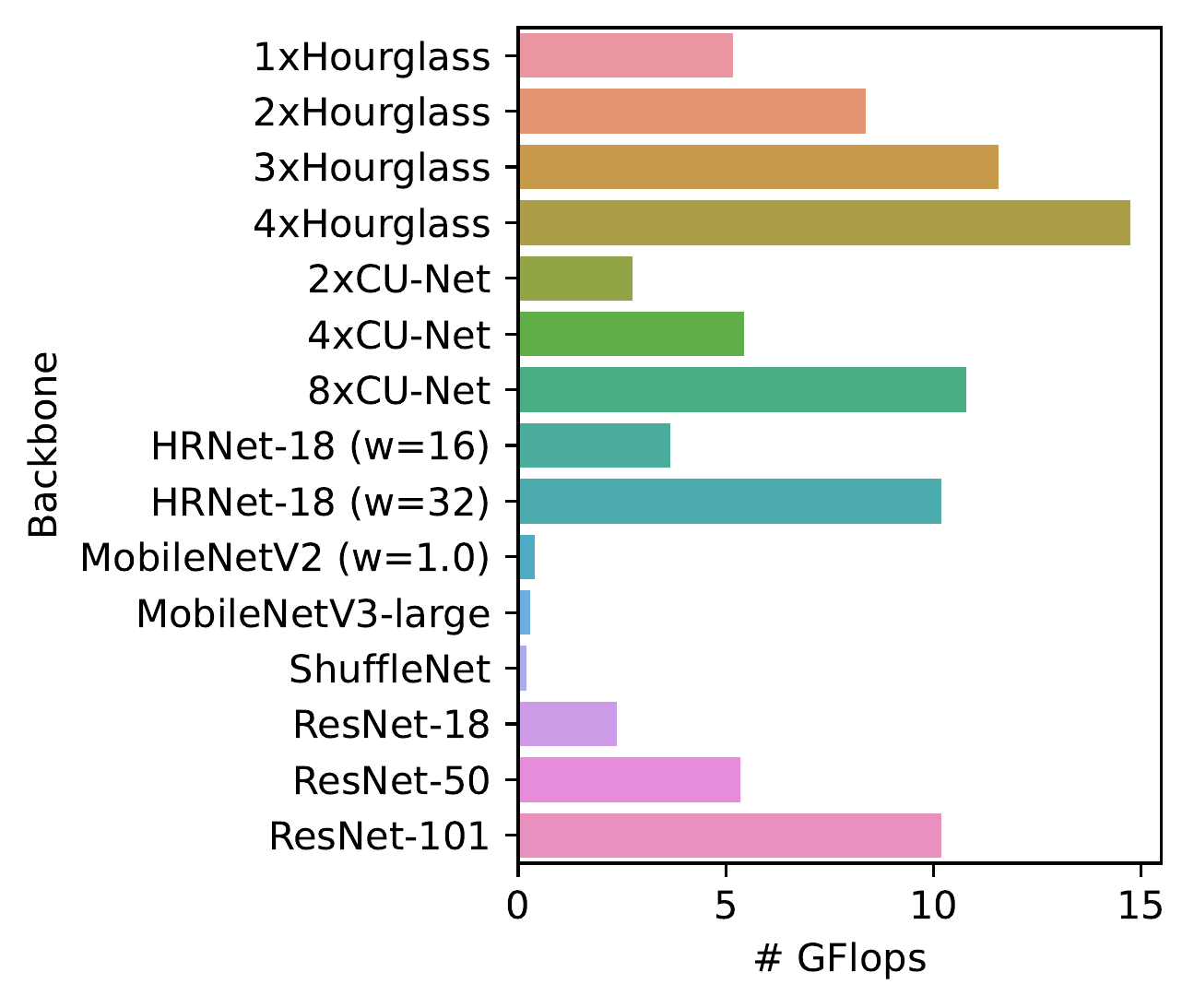}
    \caption{Number of floating-point operations}\label{fig:backbones-gflops}
  \end{subfigure}
  \hfill
  \begin{subfigure}{0.48\linewidth}
    \includegraphics[width=\linewidth]{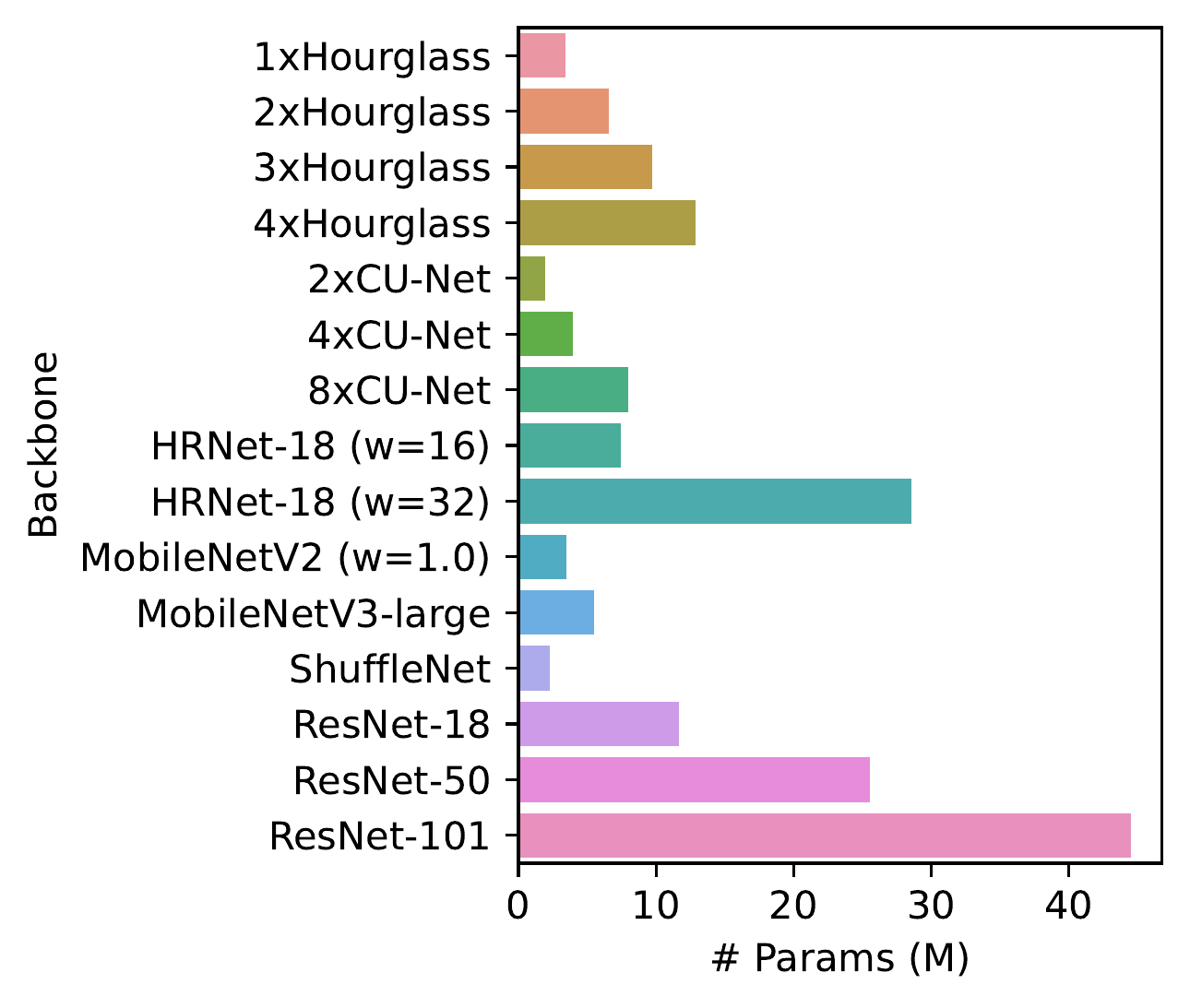}
    \caption{Number of parameters}\label{fig:backbones-params}
  \end{subfigure}
  \caption{Comparison of different backbones by the number of floating-point operations~(a) and the number of parameters~(b). Note that Hourglass, CU-Net, HRNet require much more computation, than other backbones, but have relatively small number of parameters. We use \(S \times\) to denote a stack of \(S\) modules, \(w=X\) to denote width multiplier of X.}\label{fig:backbones}
\end{figure*}

We summarize backbone performance in \cref{fig:backbones}, where we show number of floating-point operations in \cref{fig:backbones-gflops} (the greater is the number, the more time it takes to infer the network) and number of parameters in \cref{fig:backbones-params} (more parameters take more memory). Note, that Hourglass, CU-Net, HRNet require much more computation than other backbones, but have relatively small number of parameters. This is because these networks consider input at multiple resolutions and have to produce large heatmaps for each landmark. Other backbones (MobileNet, ShuffleNet and ResNet) process input at single scale and do not work with heatmaps.

Network backbones for each model considered in this survey are shown is summary \cref{sec:algorithm-summary}, \cref{tab:backbone-inference-speed}.

\subsection{Facial Landmark Detection: Novel Algorithms and Ideas}\label{sec:main-landmark-detection}

In this section we present a description of recently introduced facial landmark detection algorithms. Each description is structured as follows: backbone and algorithm type are mentioned first, followed by explanation of novel ideas and approaches. The primary focus of this paper is on modern algorithms, covering years 2018 -- 2021. We include some important older algorithm for completeness as well. The facial landmark algorithms covered in this section have been selected if: 1)~the algorithm improves state-of-the art score established in the previous year; 2)~ideas presented in the algorithm are then used in several following papers; 3)~algorithm expands applicability of facial landmark detection or presents distinctive novel idea not discussed before. If only a slight modification is presented, that doesn't improve inference speed, quality or applicability of the algorithm, such algorithm is not included in this section. The algorithms are collected from various sections, including, but not limited to, top worldwide computer vision conferences. Overall 22 algorithm discussions are presented in this section.

\begin{figure}[htb]
  \captionsetup[subfigure]{justification=centering}
  \centering
  \begin{subfigure}{0.32\linewidth}
    \includegraphics[width=0.9\linewidth]{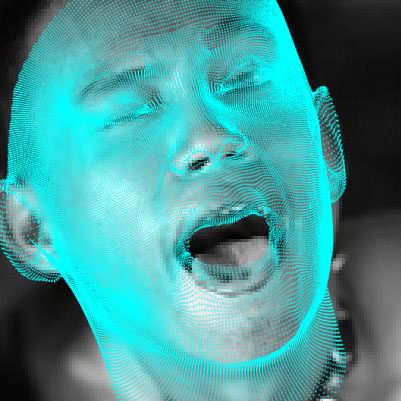}
    \caption{}
  \end{subfigure}
  \begin{subfigure}{0.32\linewidth}
    \includegraphics[width=0.9\linewidth]{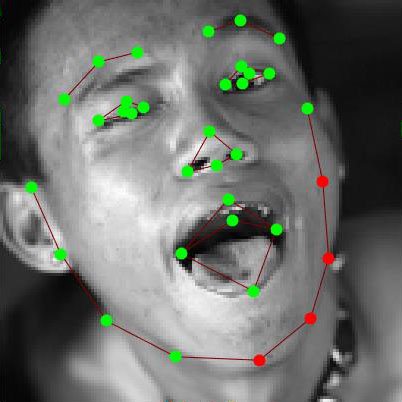}
    \caption{}
  \end{subfigure}
  \caption{(a)~DeFA dense 3D face mesh produced for different emotions and poses, (b)~DeFA facial landmark predictions acquired from the face mesh. Importantly, varied number of facial landmarks can be produced from a single face mesh.\protect\footnotemark}\label{fig:defa}
\end{figure}

\textbf{Dense Face Alignment (DeFA)}~\cite{DeFA} is a shape-model-based approach. It uses custom-built convolutional neural network as a backbone. It is the only algorithm described in this section, where a neural network is used for facial landmark prediction through a deformable 3D face mesh. Algorithm is interesting in the following: 1)~it allows to build a dense 3D face mesh using only a single 2D image. Mesh can be built for a wide range of poses and emotions (\cref{fig:defa}); 2)~DeFA can be trained jointly on datasets with different number of landmarks, as landmarks are hooked as mesh constraints.

\afterpage{
  \footnotetext{Images are included under MIT license. Source: \href{https://github.com/yaojieliu/ICCVW2017-DenseFaceAlignment}{DeFA}}
}

\textbf{Style Aggregated Network (SAN)}~\cite{SAN} is a heatmap-based approach, which is based on a modified ResNet-152 backbone. The authors have noticed style variability of photographs in 300W and AFLW datasets, which can be dark or light, colored or black \& white. Existing to date algorithms were not accounting for that information. Furthermore, the authors have noticed that depending on style, prior algorithms were predicting facial landmark locations in slightly different places, with higher error on photographs with harsh lighting conditions. As a solution they have proposed: 1)~to train Generative Adversarial Network (GAN)~\cite{GANs}, namely CycleGAN~\cite{CycleGAN} to transform images of different styles into neutral; 2)~to train another neural network to predict landmarks from two inputs: style-neural and the original image. CycleGAN colorizes grayscale images and tones down bright colors. This makes all input images have a similar color distribution, which simplifies learning of face features by a neural network. Note, that style-neutral images produced by the proposed network are not always properly colorized and might containing artifacts, because-of that the authors propose to predict landmark jointly on the original and style-neutral images.

\textbf{Look at Boundary (LAB)}~\cite{LAB} is a combined heatmap and direct regression method. A stack of 4 Hourglass modules is used to predict boundary heatmap, from which another neural network predicts landmark matrix. The key advancement of this architecture is that the authors introduce face feature boundary heatmap, which is built as an intermediate representation between original image and predicted landmarks (\cref{fig:lab-landmarks}). Such a trick improves facial landmark prediction quality. Furthermore, it allows to train boundary estimation module on several datasets with different annotation schemes at once. After boundary module, another network predicts facial landmark coordinates. It should be noted, that only boundary submodule can be trained on datasets with different annotation schemes, while the landmark regression is trained for each dataset separately. Face structural information is modeled with the use of message passing~\cite{MessagePassing1, MessagePassing2}, that is, a graph-based way to model relationships. Boundary module is trained in adversarial (GAN) fashion. As the authors have shown, pretraining the boundary module on 300W improves prediction quality on AFLW and COFW datasets. Also, a novel facial landmark dataset was introduced in the work, namely WFLW.

\afterpage{
\begin{figure}[t]
  \captionsetup[subfigure]{justification=centering}
  \centering
  \begin{subfigure}{0.32\linewidth}
    \centering
    \includegraphics[width=0.9\linewidth]{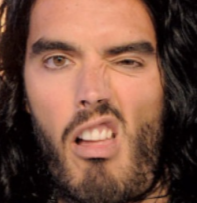}
    \caption{}
  \end{subfigure}
  \begin{subfigure}{0.32\linewidth}
    \centering
    \includegraphics[width=0.9\linewidth]{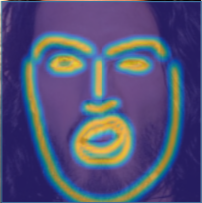}
    \caption{}
  \end{subfigure}
  \begin{subfigure}{0.32\linewidth}
    \centering
    \includegraphics[width=0.9\linewidth]{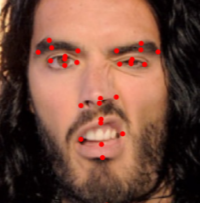}
    \caption{}
  \end{subfigure}
  \caption{LAB: (a)~image to be labeled; (b)~first module predicts intermediate boundary representation, that is common for different face landmark annotation schemes; (c)~second module predicts actual facial landmark coordinates from boundary information~\cite{LAB}.\protect\footnotemark}\label{fig:lab-landmarks}
\end{figure}
\footnotetext{Author's written consent has been acquired to include these images.}
}

\begin{figure}[htb]
  \centering
  \includegraphics[width=0.4\textwidth]{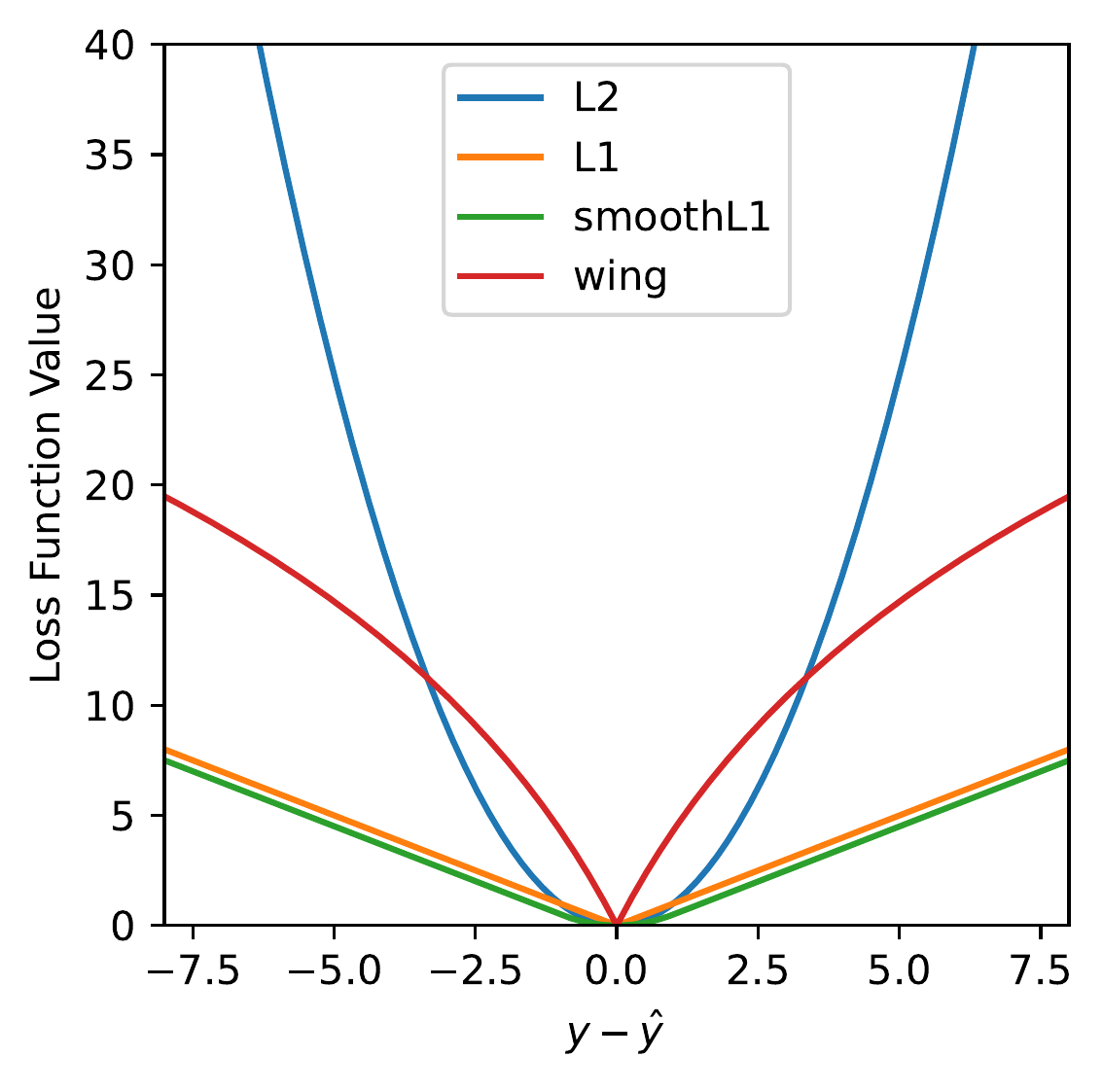}
  \caption{Loss function comparison: L2, L1, smoothL1, Wing (with \(w=15\), \(\epsilon=3\)). Note, that quadratic growth of L2 loss makes it sensitive to outliers. Thus, forcing the network to learn annotation errors. L2, L1 and smoothL1 yield very small values for small landmark location differences. This hinders network training, when network predictions are \textit{almost} correct. In contrast, Wing is less sensitive to outliers and is much sensitive to medium-to-small errors, which improves training overall.}\label{fig:wing-loss-function-comparison}
\end{figure}

\textbf{Wing Loss}~\cite{WingLoss} is a direct regression approach. Several backbones were considered: custom-built CNN-6; two-stage approach, when CNN-6 produces coarse landmarks, and CNN-7 then refines them; ResNet-50 backbone. The authors note, that the field of loss functions for facial landmark detection problem is barely studied. Most researchers use $L2 = x^2 / 2$ as a loss function for direct regression, which is known to be sensitive to outliers. For that reason, some of prior works have used $smoothL1$~\cite{FastRCNN} loss instead. The authors make a comparison of $L2$ against other loss functions, such as $L1(x) = |x|$ and $smoothL1$, which is defined as:

\begin{align}
  \label{eq:smoothL1}
  smoothL1(x)=
  \begin{cases}
    x^2 / 2,     & \text{if } x < 1  \\
    |x| - 1 / 2, & \text{otherwise},
  \end{cases}
\end{align}
and note that the latter two give better results. The main paper contribution is in introduction of a new loss, named \textit{Wing loss}, which combines $L1$ for large landmark deviations and $\ln(\cdot)$ for medium and small:

\begin{align}
  \label{eq:wing}
  wing(x)=
  \begin{cases}
    w \ln(1 + |x| / \epsilon), & \text{if } |x| < w \\
    |x| - C,                   & \text{otherwise},
  \end{cases}
\end{align}
where $C = w - w \ln(1 + w / \epsilon)$, $w$ and $\epsilon$
are hyperparameters ($w = 15$, $\epsilon = 3$ in paper).
Visual comparison of loss functions is presented in \cref{fig:wing-loss-function-comparison}.

Also, to train more on hard examples the authors introduce PDB algorithm, which works as follows: 1)~face rotation angle histogram is built; 2)~rare examples (determined via the histogram) are duplicated with augmentations. As can be seen from \cref{tab:wing-loss-aflw}, using CNN-6/7 cascade with $wing(\cdot)$ loss in combination with PDB substantially lowers the NME.

\begin{table}[htb]
  \caption{NME comparison of different loss functions on AFLW dataset. Note that Wing loss with PDB hard example mining has the best performance.}\label{tab:wing-loss-aflw}
  \centering
  \begin{tabular}{lrrrr}
    \toprule
    Network     & L2   & L1   & smoothL1 & Wing          \\
    \toprule
    CNN-6/7     & 2.06 & 1.82 & 1.84     & 1.71          \\
    CNN-6/7+PDB & 1.94 & 1.73 & 1.76     & \textbf{1.65} \\
  \end{tabular}
\end{table}

\textbf{AVS}~\cite{AVS-AFLW-68Dataset} is a heatmap-based approach. Similarly to SAN, the authors have studied style in face landmark detection. They have proposed augmenting training set by changing image style via GAN image generation. The authors have trained ResNet-18, SAN and LAB methods on their extended training set, which resulted in better performance.

\begin{figure}[htb]
  \centering
  \includegraphics[width=\linewidth]{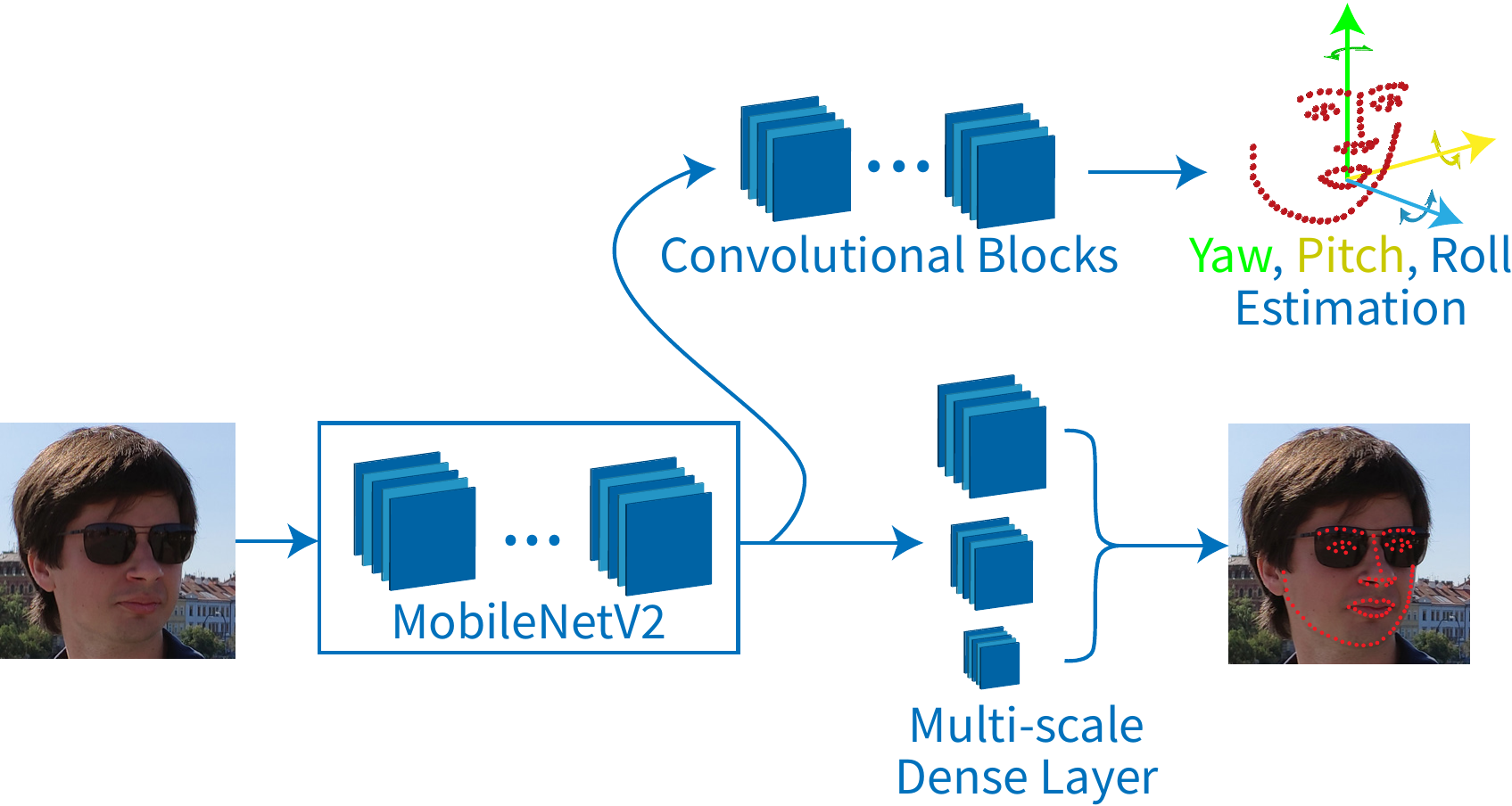}
  \caption{PFLD architecture. MobileNetV2 is used as feature extractor with multiple tasks:  1)~to predict face landmark locations multi-scale fully-connected layer is used, which better captures image features at multiple scales (lower branch); 2)~additional convolutional blocks are attached to MobileNetV2 for yaw, pitch, roll face rotation angle prediction (upper branch). Estimated angles are embedded into training loss to improve overall network performance. Estimation is not performed during network inference.}\label{fig:pfld}
\end{figure}

\textbf{Practical Facial Landmark Detector (PFLD)}~\cite{PFLD} is a direct approach. MobileNetV2 backbone with full (1X) and quarter (0.25X) width has been considered. PFLD enables fast facial landmark detection directly on a mobile device. This is, to the best of our knowledge, the only modern neural-network-based algorithm, whose authors have shown that their algorithm can work efficiently on a mobile device. MobileNetV2~\cite{MobileNetV2} is used as feature extractor in PFLD. Two heads are attached to it (\cref{fig:pfld}): 1)~facial landmark regression, where multi-scale fully-connected layer in the end of the head is used (lower branch); 2)~3D face model rotation angle estimator (yaw, pitch and roll), shown in upper branch of the figure. The second head contains a set of convolutional layers and is only used during training.

The most common datasets do not have information about 3D landmark coordinates. To get them the authors propose to 1)~build a ``mean'' face representation containing 11 facial landmarks, based on the data in the training set; 2)~estimate rotation matrix for each face between its and ``mean'' landmarks; 3)~compute yaw, pitch, roll angles from the rotation matrix. According to the authors, such an approach is not very accurate for estimating angles, yet improves network training.

Furthermore, during training, the data is weighted based on image difficulty using a special loss function:
\begin{equation}
  \label{eq:PfldLoss}
  \mathcal{L} = \frac{1}{M} \sum_{m=1}^M \sum_{n=1}^N \\
  \bigg(\sum_{c=1}^C \omega_n^c \cdot \sum_{k=1}^K (1 - \cos{\theta_n^k}) \bigg) ||d_n^m||_2^2
\end{equation}
where $N$ is the number of facial landmarks, $M$ is the number of training examples, $K=3$, $\theta_1$, $\theta_2$, $\theta_3$ are yaw, pitch, roll rotation angles of the above-described 3D face model, $d_n^m$ represents difference vector between \(n\)\textsuperscript{th} predicted and ground true facial landmarks for \(m\)\textsuperscript{th} image; $C$ is the number of complexity classes for face images (such as profile or frontal face, face-up, face-down, emotions or occlusion), $\omega_n^c$ is fraction of images in the corresponding complexity class to their total number $M$.

\textbf{FAN}~\cite{FAN} is a heatmap-based approach. A stack of 4 Hourglass modules is used. The authors modify Hourglass architecture by substituting Bottleneck block with hierarchical, parallel and multi-scale block with binary convolutions from~\cite{BinarizedConvolutionalLandmarks}. 3 methods have been trained in the work: for 2D, 3D landmark detection, and 2D-to-3D model. 2D-to-3D model serves to transform 2D landmark representation into 3D. Interestingly, the inputs to the 2D-to-3D network are image and landmark heatmaps (one for each of the input 2D landmarks). The algorithm has not been tested on conventional 2D face landmark datasets. Thus, is missing from summary in \cref{sec:algorithm-summary}. While binarized convolutions are stated to be faster, than conventional~\cite{BinarizedConvolutionalLandmarks}, no testing results have been presented. Architecture has been applied to landmark prediction in further works. Also, LS3D-W 3D face landmark dataset has been presented in this work.

\begin{figure}
  \captionsetup[subfigure]{justification=centering}
  \centering
  \includegraphics[width=0.9\linewidth]{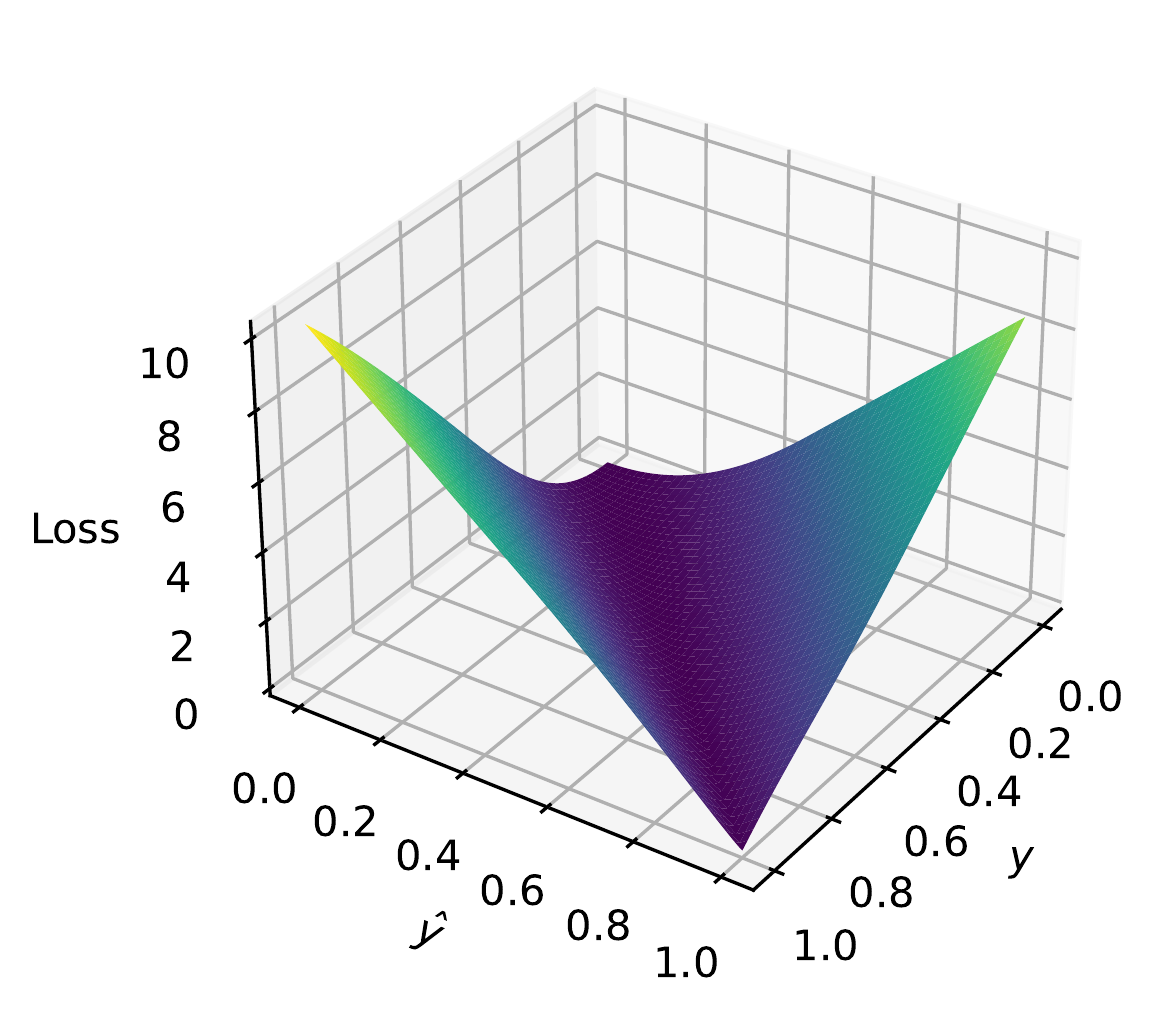}
  \caption{AWing surface plot. AWing accepts true \(y\) and predicted \(\hat{y}\) heatmap probabilities. The function behaves as L2 loss for background pixels (when \(y, \hat{y}\) are close to zero), and as Wing loss for foreground (when \(y, \hat{y}\) are close to one), while preserving continuity. Thus, a greater weight is assigned to foreground pixels, resulting in sharper heatmaps and more accurate prediction.}\label{fig:awing-surface}
\end{figure}

\textbf{AWing}~\cite{AWing} is a heatmap-based algorithm, Hourglass is used as a backbone. Stacks from 1 to 4 modules have been considered. The algorithm is based on Wing loss, FAN, LAB papers, and CoordConv~\cite{CoordConv}. The authors have noticed, that $L2$ loss function does not produce sharp-enough heatmaps on difficult face images, because it is insensitive to small errors. In the meantime, the original Wing loss is inappropriate for heatmap-based detection as its gradient is discontinuous at the point of zero. In addition, each heatmap has a class imbalance. Only a few pixels on the map relate to the foreground class (meaning that landmark is likely to be at this point), while most parts are labeled as background class. The class imbalance is also not considered in the original Wing loss implementation. To solve all these issues, Adaptive Wing loss (\cref{fig:awing-surface}) is introduced, which is 1)~differentiable around zero; 2)~accents small errors around foreground pixels, but not around background. Here we do not give the function formula due to its complexity. To predict foreground pixels even more precisely, the authors introduce a special weighted loss map, which additionally enhances sharpness of the facial landmark heatmap.

\textbf{MobileFAN}~\cite{MobileFAN} is a heatmap-based approach, based on modified MobileNetV2 backbone with 1X or 0.5X width. The authors examine network distillation approaches in order to reduce the number of model parameters and increase inference speed for heatmap-based methods. Note, that despite name similarity the approach is not based on FAN~\cite{FAN}, discussed earlier.

\textbf{Geometry Aggregated Network (GEAN)}~\cite{GEAN} is a heatmap-based approach, based on a stack of 4 Hourglass modules. The authors propose train- and test-time augmentation using Adversarial Attacks. Face adversarial attacks add noise or deformation to an image, so that face will not be recognized by face recognition system. To do that, face adversarial attack~\cite{FacesAdversarialAttack} warps the image to shift facial landmarks. The resulting face has slightly different shape, eye distance, etc. Hourglass is used to detect landmark locations on such deformed image. The detected landmarks now correspond to the warped face. However, we need to form a prediction for the original face. To do that, we shift landmark coordinates with warp deformation, that is opposite to the one introduced by the adversarial attack. Now the predicted landmarks correspond to the original face. Next, we follow this procedure for \(K\) random adversarial attacks. It turns out, that the predicted facial landmarks for each of the \(K\) images will be slightly different. Averaging such landmarks over all \(K\) images, results in accuracy improvement.

According to the authors, with respect to performance/quality ratio, it is the most beneficial to generate \(K = 5\) adversarial examples for both training and testing. It is possible to use different number of adversarial images for training and testing. The authors have explored several modifications to the adversarial attack algorithm, and the best results are obtained when attack scale is set individually for each of the landmarks' semantical groups. The groups are assigned based on face region, such as nose, eyes, eye-brows, etc.

We have a deeper look at the concept of Adversarial Attacks in \cref{sec:vulnerabilities} of this paper.

\textbf{HRNetV2}~\cite{HRNetV2} is a heatmap-based approach. In this work the original architecture of HRNet~\cite{HRNet} has been improved and applied to the task of facial landmark prediction.

\textbf{LUVLi}~\cite{LUVLi} is a heatmap-based approach. This is the only algorithm with CU-Net backbone. A stack of 8 modules is used. The authors state, that facial landmark detection is used in many critically important applications. Thus, they propose a method to predict facial landmark visibility and algorithm confidence for each landmark. Cholesky Estimator Network (CEN) and the Visibility Estimator Network (VEN) are introduced for landmark and visibility predictions correspondingly. To increase heatmap precision, the authors use weighted spatial mean of heatmap's positive elements, instead of simple argmax. Also, a relabeled AFLW dataset with 68 landmarks and landmark visibility labels is presented.

\textbf{Deep Adaptive Graph (DAG)}~\cite{DAG} is a direct regression approach. Note, that landmarks here are predicted through a graph. Multiple backbones have been considered: VGG16~\cite{VGG}, ResNet50, 4\(\times\)Hourglass, HRNet-18. HRNet-18 has shown the best results.

Face landmark prediction accuracy can be improved by taking into account structural information about human face. The authors propose a topology-adapting graph learning in a form of Graph Convolutional Network (GCN) cascade for facial and medical (e.g., hand, pelvis) landmark detection. In DAG algorithm graph \(G = (V, E, F)\) is constructed, where \(V\) is a set of vertices, \(E\) is a set of edges, \(F = \left\{f_1, f_2, \ldots, f_{|V|}\right\}\) is the so-called graph signal or graph features. Each vertex \(v\) corresponds to a single landmark. Each pair of vertices \((v_i, v_j)_{i \neq j}\) is connected via a weighted edge \(e_{ij}\). The weights \(e_{ij}\) are learned during training process, they determine how information is propagated in a graph convolution. Larger weights denote stronger semantical connection between corresponding vertices. Graph convolution is defined as follows:
\begin{equation}
  \label{eq:gcn}
  f_{k+1}^i = W_1 f_k^i + \sum_{j=1}^{|E|} {e_{ij} W_2 f_k^j}
\end{equation}
where \(W_1\) and \(W_2\) are learnable parameter matrices. \(f_k^i\) is the feature computed for \(i\)\textsuperscript{th} vertex and \(k\)\textsuperscript{th} graph convolution.

Features \(F\) contain visual \(p_i\) and shape \(q_i\) features. Visual features are acquired from feature map \(H\), that is produced by processing the whole image via convolutional neural network. Feature \(p_i\), that corresponds to \(i\)\textsuperscript{th} vertex, is then acquired from \(H\) at a location near the landmark. To obtain shape features \(q_i\) the authors compute displacement vectors for each pair of landmarks. Displacement information improves the algorithm performance, when face is partially occluded. In such cases, landmark locations can then be predicted from neighboring landmarks.

The landmark prediction process is conducted as follows:  initial graph is constructed with mean weights computed over training set. Two separate GCNs are used for iterative graph transformation. GCN-Global is used to predict perspective transformation of the initial graph. GCN-Local is then applied multiple times to predict offsets for each of the landmarks for precise graph refinement.

The authors show, that in case of significant face occlusion their algorithm is better than the competition. In addition, the learned graph is good at capturing semantical information about human face, greater weights \(e_{ij}\) are learned for landmarks that appear physically closer to each other. For instance, edges between eyebrows have greater weight than edges between eyebrow and chin landmarks.

\textbf{PropagationNet}~\cite{PropagationNet} is a heatmap-based algorithm, which uses a stack of 4 Hourglass modules as a backbone. The authors note importance of face boundary information for landmark prediction. Previous LAB approach used heavy generative adversarial network for boundary estimation. The authors of PropagationNet propose much simpler and faster approach: several convolutional blocks transform landmark heatmaps into boundary heatmaps after each Hourglass module. Boundary heatmaps serve as attention mask for intermediate predictions in Hourglass module to improve the final prediction accuracy. In addition, Hourglass module has modifications from FAN, and is extended with CoordConv~\cite{CoordConv} and Anti-aliased CNN~\cite{AntiAliasedCNN}.

Also, the authors introduce Focal Wing Loss, an extension of Adaptive Wing loss, that assigns greater weights to image scenes less presented in the current training batch. The examples of image scenes are large head pose, exaggerated expression, etc. The focus function \(\sigma_n^{(c)}\) for class \(c\) and sample \(n\) is defined as:
\begin{equation}
  \sigma_n^{(c)} =
  \begin{cases}
    1 ,                                   & \text{if }\sum_{n=1}^{N}{s_n^{(c)}} = 0 \\
    \frac{N}{\sum_{n=1}^{N}{s_n^{(c)}}} , & \text{otherwise}
  \end{cases}
\end{equation}
where \(s_n^{(c)}=0\), when the sample \(n\) does not belong to class \(c\); and \(s_n^{(c)}=1\), otherwise. Image scene annotations exist in WFLW, but not in 300W and COFW. The authors have annotated images of 300W and COFW by themselves. To form the final loss, image focus coefficient \(\sigma_n^{(c)}\) is multiplied by conventional Adaptive Wing loss.

\textbf{SAAT}~\cite{SAAT} is a heatmap-based approach, which uses a stack of 2 Hourglass modules as a backbone. The authors propose augmenting the training set with adversarial images. The network architecture is left unchanged. In contrast to GEAN, only training procedure is modified, no artificially changed images are generated at test-time. Conditional GAN is used to perform the adversarial attack.

\textbf{LDDMM-Face}~\cite{LDDMM-Face} is a shape model approach, which uses HRNet-18 as a backbone. The primary focus of the work is on cross-dataset and sparse-to-dense annotation. Sparse-to-dense means, that the network can be trained on sparse face landmarks, and then it predicts dense landmark annotations. The authors estimate shape model deformation via large deformation diffeomorphic metric mapping (LDDMM)~\cite{LDDMM-Curve, LDDMM-Landmark} method. While cross-dataset landmark annotation is out of scope of this survey, this method is also capable of achieving good results for classical face landmark datasets.

\textbf{AnchorFace}~\cite{AnchorFace} is a direct regression method. Two modified backbones have been considered: ShuffleNet-V2 (with faster inference), HRNet-18 (with better accuracy). HRNet results are present only for WFLW dataset. The authors tackle the problem of landmark prediction for images with large pose. For that they propose to configure a set of anchor templates for faces with different poses. Anchor templates are configured either manually or via KMeans clusterization on the dataset. Then the templates are refined with a network that predicts offsets and confidence of each of the anchor templates.

\textbf{PIPNet}~\cite{PIPNet} is a combined heatmap and direct regression approach. Several backbones have been considered: MobileNetV2, MobileNetV3, ResNet-18, ResNet-101, etc. In PIPNet it has been noted, that heatmap-based methods have high computational cost, but good accuracy. To alleviate the cost, the authors propose 3-head network. First head predicts coarse landmark heatmaps at lower than usual resolution. Second head predicts regression offsets. Thus, fine-tuning heatmap-based predictions. Third fine-tunes landmark predictions further by regressing offsets relative to the neighboring landmarks. All the heads share the same backbone and are computed in parallel. Additionally, ``self-learning with curriculum'' method has been introduced, where the authors try to learn on 300-W and then self-learn on other facial landmark datasets.

\begin{figure}[htb]
  \captionsetup[subfigure]{justification=centering}
  \centering
  \begin{subfigure}{0.32\linewidth}
    \centering
    \includegraphics[width=0.9\linewidth]{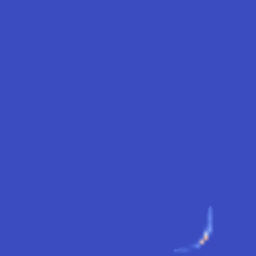}
    \caption{}\label{fig:adnet-point-edge}
  \end{subfigure}
  \begin{subfigure}{0.32\linewidth}
    \centering
    \includegraphics[width=0.9\linewidth]{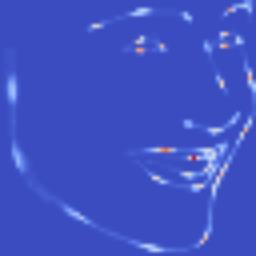}
    \caption{}\label{fig:adnet-point-edge-sum}
  \end{subfigure}
  \caption{ADNet learns additional semantical information about face in a form of Point-Edge heatmaps. Each of the heatmaps corresponds to face landmark and a part of face boundary~(a). Sum of all Point-Edge heatmaps forms face boundary~(b). Such heatmaps are predicted after each Hourglass module and are used as attention masks to improve overall network performance.\protect\footnotemark}\label{fig:adnet}
\end{figure}

\textbf{ADNet}~\cite{ADNet} is a heatmap-based approach. It uses a stack of 4 Hourglass modules as a backbone. The work is based on LAB and PropagationNet. Facial landmark datasets are annotated by humans. Thus, there exists some annotation error. It turns out, that error in tangent direction (relative to face boundary) is much higher, than in normal direction. Loss functions of existing algorithms do not account for such difference in annotation error. To mitigate the problem, anisotropic direction loss (ADL) is introduced, where higher weight is assigned to normal error. Also, Point-Edge heatmaps are presented, that are used as attention mask. Point-Edge heatmaps are predicted after each Hourglass module and have greater than zero values around face boundary corresponding to a landmark. This is shown in ~\cref{fig:adnet-point-edge}. Note landmark center accentuation shown in red. Sum of all Point-Edge heatmaps forms face boundary information (\cref{fig:adnet-point-edge-sum}). Landmark locations are obtained through soft-argmax. The final loss function that is used to train the model consists of a sum of AWing loss for Point and Edge heatmaps, as well as ADL loss for landmark heatmaps.

\footnotetext{Based on \href{https://arxiv.org/pdf/2109.05721.pdf}{this source}. Contains representative samples of Point-Edge heatmaps. Distributed under CC BY-NC-SA 4.0 License.}

\textbf{HIH}~\cite{HIH} and \textbf{SubpixelHeatmap}~\cite{SubpixelHeatmap} are heatmap-based approaches. Both use a stack of 2 Hourglass modules as backbones. The algorithms focus on reducing heatmap quantization error. Input image and landmark annotations are of resolution \(256 \times 256\). However, the heatmap is typically of size \(64 \times 64\) per landmark, which is \(1/16\)\textsuperscript{th} of the source image resolution. The landmark location is then found using argmax. The process of mapping floating point landmark location to a discrete grid is called quantization.

\textbf{HIH}. The authors tackle the problem by splitting heatmaps into \textit{integer} and \textit{decimal}. Integer heatmap is predicted via the usual heatmap-based facial landmark prediction pipeline. Then another decimal heatmap block predicts a precise offset to the quantized landmark locations. Two ways to predict the offset have been considered: based on Convolutional Neural Networks and Transformers~\cite{AttentionIsAllYouNeed}, denoted as \(\text{HIH}_C\) and \(\text{HIH}_T\) correspondingly.

\textbf{SubpixelHeatmap}. To alleviate the above-described quantization problem, the authors propose a different approach, namely local soft-argmax computation. For a given heatmap \(H_k\), \(k\)\textsuperscript{th} landmark location is first found as usual via: \((\hat{y}_k^{[1]},\hat{y}_k^{[2]}) = \arg \max H_k\) and then refined via local soft-argmax over the neighboring patch \(d\times d\):
\begin{equation}
  \label{eq:local-soft-argmax}
  (\Delta\hat{y}_k^{[1]},\Delta\hat{y}_k^{[2]}) = \sum_{m,n}\texttt{softmax}(\tau H_k)_{m,n}(m,n),
\end{equation}
where \(\tau = 10\) is the temperature, \(d = 5\) is the suggested region size. Then the final landmark location is found using: \((\hat{y}_k^{[1]} + \Delta\hat{y}_k^{[1]}-l, \hat{y}_k^{[2]} + \Delta\hat{y}_k^{[2]} -l)\), where \(\hat{y}_k^{[1]},\hat{y}_k^{[2]}\) denote landmark position on X and Y axes correspondingly, and \(l = d/2\). The authors compare this approach to global soft-argmax (i.e., computed over the whole image) applied to pose estimation in~\cite{GlobalSoftArgmax}, and state that local soft-argmax yields much better results.

Also, the authors apply test-time augmentation to improve network performance. They feed 2 images with different random augmentations (\(T_0, T_1\)) through the network \(\Phi\) aggregating the final heatmap as follows: \(H = T_0^{-1}(\Phi(T_0(\mathbf{X}_i), \theta)) + T_1^{-1}(\Phi(T_1(\mathbf{X}_i)), \theta)\), where \(\theta\) is the network \(\Phi\) parameter matrix. In addition, Hourglass architecture has been modified following FAN algorithm.

\subsection{Facial Landmark Detection Algorithms: Summary}\label{sec:algorithm-summary}

In this section we present and discuss facial landmark detection algorithm performance on the most widely used datasets: 300W, AFLW, COFW and WFLW. We summarize backbones used, and inference times on desktop and mobile devices. We present brief summary of contributions of each facial landmark detection method. We conclude this section with analysis of algorithm performance by years and per algorithm type.

~\cref{tab:300w-inter-ocular,tab:300w-inter-pupil,tab:aflw,tab:cofw,tab:wflw} present facial landmark detection method metrics on the most common datasets. The the best result is shown in red, second best is shown in blue. Metrics in the tables include NME (\%), Failure Rate (FR, \%) and CED-AUC. \cref{tab:backbone-inference-speed} has method backbones and inference times listed. Different hardware was used for algorithm inference speed measurements. So instead of defining first and second fastest, we show all algorithms that perform faster than 60 frames per second (17~ms) in green. Note, that in addition to face detection, other algorithms will need to be executed, that is why the threshold is so strict. The tables are filled based on the results presented in the corresponding papers. If the result was published later, the metric's source is shown in square brackets. \cref{tab:landmark-detection-method-summary} has a brief summary of algorithm novelties proposed in each paper.

We present 300W dataset results normalized by both inter-ocular distance in \cref{tab:300w-inter-ocular} and inter-pupil in \cref{tab:300w-inter-pupil}. Metrics are split into common, challenge, and full as per protocol. Can be noticed, most novel algorithms use inter-ocular distance normalization. As is shown in \cref{tab:300w-inter-ocular}, error on challenging subset is still quite high (3.99~\%) and is significantly higher than the best common subset error (2.53~\%). From \cref{tab:300w-inter-pupil} we note that Wing neural-network-based algorithm with ResNet-50 backbone is 1.7 times better, than regression-tree-based ERT.

\begin{table*}[tp]
  \caption{Face landmark detection normalized mean error (NME) on 300-W dataset. Inter-ocular normalization is used. The best result is shown in red, second best in blue. Note, that significant qualitative improvement has been achieved over the past few years, but still Challenge subset error is quite high.}\label{tab:300w-inter-ocular}
  \centering
  \begin{tabular}{lcccc}
    \toprule
    Model                                  & Year                   & Common                 & Challenge              & Full                  \\
    \toprule

    DeFA~\cite{DeFA}                       & 2017                   & 5.37                   & 9.38                   & 6.10                  \\
    \hline

    SAN~\cite{SAN}                         & \multirow{2}{*}{2018}  & 3.34                   & 6.60                   & 3.98                  \\
    LAB~\cite{LAB}                         &                        & 2.98                   & 5.19                   & 3.49                  \\
    \hline

    AVS~\cite{AVS-AFLW-68Dataset}          & \multirow{8}{*}{2019}  & 3.21                   & 6.49                   & 3.86                  \\
    PFLD 0.25X~\cite{PFLD}                 &                        & 3.03                   & 5.15                   & 3.45                  \\
    PFLD 1X                                &                        & 3.01                   & 5.08                   & 3.40                  \\
    PFLD 1X+ (extra data)                  &                        & 2.96                   & 4.98                   & 3.37                  \\
    AWing-1HG~\cite{AWing}                 &                        & 2.81                   & 4.72                   & 3.18                  \\
    AWing-2HG                              &                        & 2.77                   & 4.58                   & 3.12                  \\
    AWing-3HG                              &                        & 2.73                   & 4.58                   & 3.10                  \\
    AWing                                  &                        & 2.72                   & 4.52                   & 3.07                  \\
    \hline

    MobileFAN (0.5)~\cite{MobileFAN}       & \multirow{7}{*}{2020}  & 4.22                   & 6.87                   & 4.74                  \\
    MobileFAN                              &                        & 2.98                   & 5.34                   & 3.45                  \\
    GEAN (extra data)~\cite{GEAN}          &                        & 2.68                   & 4.71                   & 3.05                  \\
    HRNetV2~\cite{HRNetV2}                 &                        & 2.87                   & 5.15                   & 3.32                  \\
    LUVLi~\cite{LUVLi}                     &                        & 2.76                   & 5.16                   & 3.23                  \\
    DAG~\cite{DAG}                         &                        & 2.62                   & 4.77                   & 3.04                  \\
    PropagationNet~\cite{PropagationNet}   &                        & 2.67                   & \textcolor{red}{3.99}  & \textcolor{red}{2.93} \\
    \hline

    SAAT~\cite{SAAT}                       & \multirow{11}{*}{2021} & 2.87                   & 5.03                   & 3.29                  \\
    LDDMM-Face~\cite{LDDMM-Face}           &                        & 3.07                   & 5.40                   & 3.53                  \\
    AnchorFace~\cite{AnchorFace}           &                        & 3.12                   & 6.19                   & 3.72                  \\
    PIPNet (MobileNetV2)~\cite{PIPNet}     &                        & 2.94                   & 5.30                   & 3.40                  \\
    PIPNet (MobileNetV3)                   &                        & 2.94                   & 5.07                   & 3.36                  \\
    PIPNet (ResNet-18)                     &                        & 2.91                   & 5.18                   & 3.36                  \\
    PIPNet (ResNet-101)                    &                        & 2.78                   & 4.89                   & 3.19                  \\
    ADNet~\cite{ADNet}                     &                        & \textcolor{red}{2.53}  & 4.58                   & \textcolor{red}{2.93} \\
    \(\text{HIH}_C\)~\cite{HIH}            &                        & 2.95                   & 5.04                   & 3.36                  \\
    \(\text{HIH}_T\)                       &                        & 2.93                   & 5.00                   & 3.33                  \\
    SubpixelHeatmap~\cite{SubpixelHeatmap} &                        & \textcolor{blue}{2.61} & \textcolor{blue}{4.13} & 2.94                  \\
  \end{tabular}
\end{table*}

\begin{table*}[tp]
  \caption{Face landmark detection normalized mean error (NME) on 300-W dataset. Inter-pupil normalization is used. The best result is shown in red, second best in blue. Note substantial error decrease of recently introduced neural-network-based approaches over regression tree-based (ERT) algorithm.}\label{tab:300w-inter-pupil}
  \centering
  \begin{tabular}{lcccc}
    \toprule
    Model                                & Year                  & Common                 & Challenge              & Full                   \\
    \toprule

    ERT~\cite{ERT}                       & 2014                  & -                      & -                      & 6.40~\cite{WingLoss}   \\
    \hline

    LAB~\cite{LAB}                       & \multirow{4}{*}{2018} & 3.42                   & 6.98                   & 4.12                   \\
    Wing (CNN-6)~\cite{WingLoss}         &                       & 3.35                   & 7.20                   & 4.10                   \\
    Wing (CNN-6/7)                       &                       & \textcolor{blue}{3.27} & 7.18                   & 4.04                   \\
    Wing (ResNet-50)                     &                       & \textcolor{red}{3.01}  & \textcolor{blue}{6.01} & \textcolor{red}{3.60}  \\
    \hline

    PFLD 0.25X~\cite{PFLD}               & \multirow{4}{*}{2019} & 3.38                   & 6.83                   & 4.02                   \\
    PFLD 1X                              &                       & 3.32                   & 6.56                   & \textcolor{blue}{3.95} \\
    PFLD 1X+ (extra data)                &                       & 3.17                   & 6.33                   & 3.76                   \\
    AWing~\cite{AWing}                   &                       & 3.77                   & 6.52                   & 4.31                   \\
    \hline
    DAG~\cite{DAG}                       & \multirow{2}{*}{2020} & 3.64                   & 6.88                   & 4.27                   \\
    PropagationNet~\cite{PropagationNet} &                       & 3.70                   & \textcolor{red}{5.75}  & 4.10                   \\
    \hline

    ADNet~\cite{ADNet}                   & 2021                  & 3.51                   & 6.47                   & 4.08                   \\
  \end{tabular}
\end{table*}

AFLW results are shown in \cref{tab:aflw}. NME (\%) normalized by face bounding box diagonal is used to present the results. Errors are on average smaller than on 300W possibly as AFLW has fewer landmark to be annotated (21 vs 68 in 300W), and due to different normalization. Face diagonal is larger than inter-ocular distance.

\begin{table}[htb]
  \caption{Face landmark detection normalized mean error (NME) on AFLW. Normalization by face bounding box diagonal is used. The best result is shown in red, second best in blue.}\label{tab:aflw}
  \centering
  \begin{tabular}{lcc}
    \toprule
    Model                & Full                   & Frontal                \\
    \toprule
    ERT                  & 4.35~\cite{SAN}        & 2.75~\cite{SAN}        \\
    SAN                  & 1.91                   & 1.85                   \\
    LAB                  & 1.85                   & 1.62                   \\
    LAB (extra data)     & 1.25                   & 1.14                   \\
    Wing (CNN-6)         & 1.83                   & -                      \\
    Wing (CNN-6/7)       & 1.65                   & -                      \\
    Wing (ResNet-50)     & 1.47                   & -                      \\
    PFLD 0.25X           & 2.07                   & -                      \\
    PFLD 1X              & 1.88                   & -                      \\
    AWing                & 1.53                   & 1.38                   \\
    GEAN (extra data)    & 1.59                   & 1.34                   \\
    HRNetV2              & 1.57                   & 1.46                   \\
    LUVLi                & \textcolor{blue}{1.39} & \textcolor{blue}{1.19} \\
    AnchorFace           & 1.56                   & 1.38                   \\
    PIPNet (MobileNetV2) & 1.52                   & -                      \\
    PIPNet (MobileNetV3) & 1.52                   & -                      \\
    PIPNet (ResNet-18)   & 1.48                   & -                      \\
    PIPNet (ResNet-101)  & 1.42                   & -                      \\
    SubpixelHeatmap      & \textcolor{red}{1.31}  & \textcolor{red}{1.12}  \\
  \end{tabular}
\end{table}

COFW results are presented in \cref{tab:cofw}. The results in papers are presented either normalized by inter-pupil or inter-ocular (majority) distance, but not both, which makes direct comparison more difficult. NME~(\%) and Failure Rate~(FR,~\%) are used to present the results. Interestingly, novel approaches have FR equal to 0.0, which means that no images in the test set have NME above 10\% as follows from Eq.~\ref{eq:FailureRate}.

\begin{table}[htb]
  \caption{Face landmark detection normalized mean error (NME) and failure rate (FR) on COFW. The best result is shown in red, second best in blue.}\label{tab:cofw}
  \centering
  \begin{tabular}{lll}
    \toprule
    Model                & NME (\%)               & FR (\%)                \\
    \toprule
    \multicolumn{3}{c}{Inter-pupil normalization}                          \\
    Wing                 & 5.44~\cite{AWing}      & 3.75~\cite{AWing}      \\
    AWing                & 4.94                   & 0.99                   \\
    PropagationNet       & \textcolor{red}{3.71}  & \textcolor{red}{0.20}  \\
    ADNet                & \textcolor{blue}{4.68} & \textcolor{blue}{0.59} \\
    \midrule
    \multicolumn{3}{c}{Inter-ocular normalization}                         \\
    LAB                  & 5.58                   & 2.76                   \\
    LAB (extra data)     & 3.92                   & 0.39                   \\
    Wing (ResNet-50)     & 5.07~\cite{PIPNet}     & -                      \\
    MobileFAN (0.5)      & 3.68                   & 0.59                   \\
    MobileFAN            & 3.66                   & 0.59                   \\
    HRNetV2              & 3.45                   & 0.19                   \\
    PIPNet (MobileNetV2) & 3.43                   & -                      \\
    PIPNet (MobileNetV3) & 3.40                   & -                      \\
    PIPNet (ResNet-18)   & 3.31                   & -                      \\
    PIPNet (ResNet-101)  & \textcolor{blue}{3.08} & -                      \\
    \(\text{HIH}_C\)     & 3.29                   & \textcolor{red}{0.0}   \\
    \(\text{HIH}_T\)     & 3.28                   & \textcolor{red}{0.0}   \\
    SubpixelHeatmap      & \textcolor{red}{3.02}  & \textcolor{red}{0.0}   \\
  \end{tabular}
\end{table}

\begin{figure}[ht]
  \centering
  \includegraphics[width=\linewidth]{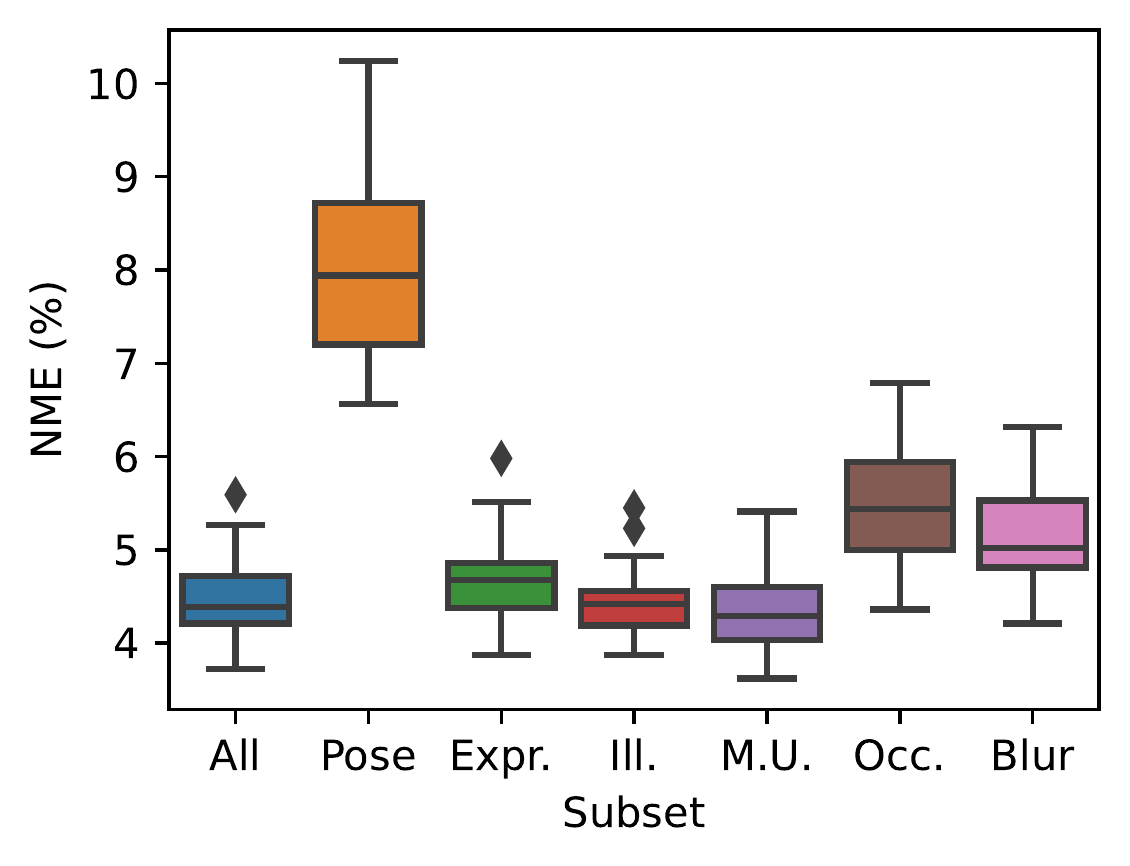}
  \caption{Box plots that show Normalized Mean Error (NME, \%) on WFLW dataset for algorithms from \cref{tab:wflw}. The results are shown for whole test set (All), as well as for subsets that focus on unusual Pose, Expression (Expr.), Illumination (Ill.), Make-Up (M.U.), Occlusion (Occ.) and Blur. Images with large pose, occlusion and blur are more challenging, than others.}\label{fig:wflw-subsets-nme}
\end{figure}

WFLW is the most recent and interesting dataset in this survey. Results are presented in \cref{tab:wflw}. We show NME~(\%), failure rate~(FR,~\%) and CED-AUC (denoted as AUC in the table) for the whole test set. Note, that lower values of NME and FR are better. In contrast, higher AUC values are better. We also present errors on all types of challenging image categories present in WFLW dataset: Pose, Expression (Expr.), Illumination (Ill.), Make-Up (M.U.), Occlusion (Occ.) and Blur. In \cref{fig:wflw-subsets-nme} box plots for different image categories are shown. The plot is based on NME for all algorithms present in \cref{tab:wflw}. We note significant difference in NME for different subsets. The most significant challenge to the landmark detection datasets comes from large pose (best error is still at 6.56~\%), followed by occlusion (4.36~\%) and blur (4.21~\%). In contrast, from make-up (3.62~\%), illumination (3.87~\%) and expression (3.87~\%) comes the least challenge. Unlike COFW, failure rate is still quite high (1.55~\%) on the test set. While a factor of 1.6 improvement has been achieved on large pose subset over the past 3 years, further improvement is welcome. We see this dataset as the one posing the most interest for novel research.

\begin{table*}[t]
  \caption{Face landmark detection normalized mean error (NME), failure rate (FR), CED-AUC on WFLW. Normalization by inter-ocular distance is used. Results are presented both for the whole test set and for subsets that focus on unusual Pose, Expression (Expr.), Illumination (Ill.), Make-Up (M.U.), Occlusion (Occ.) and Blur. The best result is shown in red, second best in blue.}\label{tab:wflw}
  \centering
  \begin{tabular}{llllllllll}
    \toprule
                                  & \multicolumn{3}{c}{Test set} & \multicolumn{6}{c}{Subsets (NME \%, \(\downarrow\))}                                                                                                                                                                                  \\
    Model                         & NME \%, \(\downarrow\)       & FR \%, \(\downarrow\)                                & AUC  \(\uparrow\)        & Pose                   & Expr.                  & Ill.                   & M.U.                   & Occ.                   & Blur                   \\
    \toprule
    LAB                           & 5.27                         & 7.56                                                 & 0.5323                   & 10.24                  & 5.51                   & 5.23                   & 5.15                   & 6.79                   & 6.32                   \\
    Wing (tested in~\cite{AWing}) & 5.11                         & 6.00                                                 & 0.5504                   & 8.75                   & 5.36                   & 4.93                   & 5.41                   & 6.37                   & 5.81                   \\
    AVS                           & 4.39                         & 4.08                                                 & 0.5913                   & 8.42                   & 4.68                   & 4.24                   & 4.37                   & 5.60                   & 4.86                   \\
    AWing                         & 4.36                         & 2.84                                                 & 0.5719                   & 7.38                   & 4.58                   & 4.32                   & 4.27                   & 5.19                   & 4.96                   \\
    MobileFAN (0.5)               & 5.59                         & 6.72                                                 & 0.4682                   & 9.68                   & 5.98                   & 5.45                   & 5.33                   & 6.49                   & 6.31                   \\
    MobileFAN                     & 4.93                         & 5.32                                                 & 0.5296                   & 8.72                   & 5.27                   & 4.93                   & 4.70                   & 5.94                   & 5.73                   \\
    HRNetV2                       & 4.60                         & -                                                    & -                        & 7.94                   & 4.85                   & 4.55                   & 4.29                   & 5.44                   & 5.42                   \\
    LUVLi                         & 4.37                         & 3.12                                                 & 0.577                    & -                      & -                      & -                      & -                      & -                      & -                      \\
    DAG                           & 4.21                         & 3.04                                                 & 0.5893                   & 7.36                   & 4.49                   & 4.12                   & 4.05                   & 4.98                   & 4.82                   \\
    PropagationNet                & 4.05                         & 2.96                                                 & 0.6158                   & 6.92                   & \textcolor{red}{3.87}  & \textcolor{blue}{4.07} & \textcolor{blue}{3.76} & \textcolor{blue}{4.58} & \textcolor{blue}{4.36} \\
    LDDMM-Face                    & 4.63                         & 3.68                                                 & 0.5509                   & -                      & -                      & -                      & -                      & -                      & -                      \\
    AnchorFace                    & 4.62                         & 4.20                                                 & 0.5516                   & -                      & -                      & -                      & -                      & -                      & -                      \\
    AnchorFace (HRNet-18)         & 4.32                         & 2.96                                                 & 0.5769                   & -                      & -                      & -                      & -                      & -                      & -                      \\
    SAAT                          & 5.11                         & 5.63                                                 & 0.5633                   & -                      & -                      & -                      & -                      & -                      & -                      \\
    PIPNet (MobileNetV2)          & 4.79                         & -                                                    & -                        & 8.76                   & 4.86                   & 4.56                   & 4.60                   & 6.04                   & 5.53                   \\
    PIPNet (MobileNetV3)          & 4.65                         & -                                                    & -                        & 8.22                   & 4.75                   & 4.49                   & 4.46                   & 5.72                   & 5.31                   \\
    PIPNet (ResNet-18)            & 4.57                         & -                                                    & -                        & 8.02                   & 4.73                   & 4.39                   & 4.38                   & 5.66                   & 5.25                   \\
    PIPNet (ResNet-101)           & 4.31                         & -                                                    & -                        & 7.51                   & 4.44                   & 4.19                   & 4.02                   & 5.36                   & 5.02                   \\
    ADNet                         & 4.14                         & 2.72                                                 & 0.6022                   & \textcolor{blue}{6.96} & 4.38                   & 4.09                   & 4.05                   & 5.06                   & 4.79                   \\
    ADNet (focal loss)            & \textcolor{blue}{3.98}       & \textcolor{blue}{2.00}                               & \textcolor{blue}{0.6250} & \textcolor{red}{6.56}  & \textcolor{blue}{4.02} & \textcolor{red}{3.87}  & \textcolor{red}{3.62}  & \textcolor{red}{4.36}  & \textcolor{red}{4.21}  \\
    \(\text{HIH}_C\)              & 4.18                         & 2.96                                                 & 0.597                    & 7.20                   & 4.19                   & 4.45                   & 3.97                   & 5.00                   & 4.81                   \\
    \(\text{HIH}_T\)              & 4.21                         & 2.84                                                 & 0.593                    & 7.20                   & 4.28                   & 4.42                   & 4.03                   & 5.00                   & 4.79                   \\
    SubpixelHeatmap               & \textcolor{red}{3.72}        & \textcolor{red}{1.55}                                & \textcolor{red}{0.631}   & -                      & -                      & -                      & -                      & -                      & -                      \\
  \end{tabular}
\end{table*}

Due to complexity of manual dense facial landmark annotation, the datasets are quite small. Thus, additional training data has a significant impact on model accuracy. We group additional data used into 3 main groups: 1)~backbone pretraining on ImageNet; 2)~usage of extra image labels (such as image scene); 3)~pretraining on other face-related datasets. SAN, DAG, PIPNet, Wing~(ResNet-50 only) state that they use ImageNet-pretrained backbones. Hence, they are related to the first group. Second group with PFLD, PropagationNet, annotates images manually with categories (i.e., significant pose, emotion) to assign higher weights to rare categories. Also, PropagationNet and ADNet (focal loss modification only) use weighted loss based on image classes directly available from WFLW dataset. The final third group of algorithms uses extra face-related data during training. GEAN uses pretrained face recognizer to perform an adversarial attack on; certain modifications of LAB use pretrained boundary module on 300W dataset and report results on COFW and AFLW; PFLD (1X+ modification only) is pretrained on WFLW and then reported on 300W. We denote face-data-based pretraining (3\textsuperscript{rd} category) with \textit{extra data} label. We do not highlight such results as the best result; however, the data is still present in~\cref{tab:300w-inter-ocular,tab:300w-inter-pupil,tab:aflw,tab:cofw,tab:wflw}. Note the significant positive impact of LAB boundary module pretraining for AFLW and COFW dataset performance in \cref{tab:aflw,tab:cofw} correspondingly.

In \cref{tab:backbone-inference-speed} we present algorithm backbones, number of network parameters, floating-point operations, and inference times on desktop computers (CPU, GPU) and mobile phones. Hourglass and CU-Net backbones are typically stacked. We use \(N\times\)Hourglass to denote a stack of \(N\) Hourglass modules. Number of parameters translates to device memory consumption, which is especially important for mobile and edge devices. Gigaflops (GFlops) is a number of floating-point operations needed for network inference, which determines a requirement for device performance. We also present an estimate of algorithms' inference time on desktop Central Processing Unit (CPU), Graphical Processing Unit (GPU) and mobile phones, as measured by the authors themselves. Note that different hardware has been used for the experiments, affecting measurements. Unfortunately, most of the algorithms report only GPU inference speed, and 11 out of 22 reviewed algorithms do not report any speed measurements at all. While state-of-the-art approaches seem to be very computationally intensive, there are still lightweight models that are quite accurate. Hourglass backbone is the most pervasive across modern landmark detection methods (used in 9 out of 22 cases). Only one algorithm (PFLD) has been adapted to a mobile device and can run there at real-time speed. We expect several of the fastest algorithms, like Wing~(CNN6), MobileFAN~(0.5), PIPNet~(ResNet-18), and possibly AWing (1\(\times\)Hourglass) to be applicable to mobile devices as well. In general, we would like to see a larger number of fast approaches in future.

\begin{table*}[t]
  \caption{Comparison of neural network backbones, computational complexity and inference speed of facial landmark detection algorithms. The smallest number of parameters and floating-point operations (flops) is shown in red, second best in blue. Inference times of less than 17 ms (or more than 60 frames per second) are shown in green.}\label{tab:backbone-inference-speed}
  \centering
  \begin{tabular}{llrrrrr}
    \toprule
    Model            & Backbone             & \# Params (M)               & \# GFlops                   & CPU (ms)               & GPU (ms)                & Mobile (ms)            \\
    \toprule
    ERT              & -                    & -                           & -                           & \textcolor{green}{1}   & -                       & -                      \\
    DeFA             & CNN                  & -                           & -                           & -                      & -                       & -                      \\
    SAN              & ResNet-152           & -                           & -                           & -                      & 343~\cite{PFLD}         & -                      \\
    LAB              & 4\(\times\)Hourglass & 25.1~\cite{MobileFAN}       & 18.85~\cite{PropagationNet} & 2600~\cite{PFLD}       & 60                      & -                      \\
    Wing             & CNN-6                & 3.8                         & -                           & \textcolor{green}{6.7} & \textcolor{green}{2.5}  & -                      \\
    Wing             & CNN-6/7              & 12.3                        & -                           & 50                     & \textcolor{green}{5.9}  & -                      \\
    Wing             & ResNet-50            & 25                          & 5.5~\cite{PIPNet}           & 125                    & 33.3                    & -                      \\
    AVS              & ResNet-152           & 35.02~\cite{HIH}            & 33.87~\cite{HIH}            & -                      & -                       & -                      \\
    PFLD 0.25X       & MobileNetV2          & -                           & -                           & \textcolor{green}{1.2} & \textcolor{green}{1.2}  & \textcolor{green}{7.0} \\
    PFLD 1X/1X+      & MobileNetV2          & -                           & -                           & \textcolor{green}{6.1} & \textcolor{green}{3.5}  & 26.4                   \\
    FAN              & 4\(\times\)Hourglass & 24                          & -                           & -                      & 33.3                    & -                      \\
    AWing-1HG        & 1\(\times\)Hourglass & -                           & -                           & -                      & \textcolor{green}{8.3}  & -                      \\
    AWing-2HG        & 2\(\times\)Hourglass & -                           & -                           & -                      & \textcolor{green}{15.7} & -                      \\
    AWing-3HG        & 3\(\times\)Hourglass & -                           & -                           & -                      & 22.1                    & -                      \\
    AWing            & 4\(\times\)Hourglass & 24.15~\cite{PropagationNet} & 26.79~\cite{AnchorFace}     & -                      & 29.0                    & -                      \\
    MobileFAN (0.5)  & MobileNetV2          & \textcolor{red}{1.84}       & \textcolor{blue}{0.45}      & -                      & \textcolor{green}{4.0}  & -                      \\
    MobileFAN        & MobileNetV2          & \textcolor{blue}{2.02}      & 0.72                        & -                      & \textcolor{green}{4.2}  & -                      \\
    GEAN             & 4\(\times\)Hourglass & -                           & -                           & -                      & 58.8                    & -                      \\
    HRNetV2          & HRNet-18             & 9.3                         & 4.3                         & -                      & -                       & -                      \\
    LUVLi            & 8\(\times\)CU-Net    & -                           & -                           & -                      & \textcolor{green}{17}   & -                      \\
    DAG              & HRNet-18             & -                           & -                           & -                      & -                       & -                      \\
    PropagationNet   & 4\(\times\)Hourglass & 36.30                       & 42.83                       & -                      & -                       & -                      \\
    SAAT             & 2\(\times\)Hourglass & -                           & -                           & -                      & -                       & -                      \\
    LDDMM-Face       & HRNet-18             & -                           & -                           & -                      & -                       & -                      \\
    AnchorFace       & ShuffleNet-V2        & -                           & 1.71                        & -                      & 22.2                    & -                      \\
    AnchorFace       & HRNet-18             & -                           & 5.30                        & -                      & -                       & -                      \\
    PIPNet           & MobileNetV2          & 4.2                         & 0.5                         & 33.9                   & \textcolor{green}{8.3}  & -                      \\
    PIPNet           & MobileNetV3          & 4.5                         & \textcolor{red}{0.4}        & 35.2                   & \textcolor{green}{12.5} & -                      \\
    PIPNet           & ResNet-18            & 12.0                        & 2.4                         & 28.0                   & \textcolor{green}{5.0}  & -                      \\
    PIPNet           & ResNet-101           & 45.7                        & 10.5                        & 113.6                  & 17.9                    & -                      \\
    ADNet            & 4\(\times\)Hourglass & 13.37                       & 17.04                       & -                      & 95.29                   & -                      \\
    \(\text{HIH}_C\) & 2\(\times\)Hourglass & 14.47                       & 10.38                       & -                      & -                       & -                      \\
    \(\text{HIH}_T\) & 2\(\times\)Hourglass & 28.18                       & 10.29                       & -                      & -                       & -                      \\
    SubpixelHeatmap  & 2\(\times\)Hourglass & -                           & -                           & -                      & -                       & -                      \\
  \end{tabular}
\end{table*}

The final \cref{tab:landmark-detection-method-summary} presents a summary of facial landmark detection methods, where we show algorithm type, main contribution and notes on algorithm applicability and performance. We use the following abbreviations for algorithm type: D is direct regression, H is heatmap-based regression, SM is shape model, H~+~D indicates combined methods that use both heatmap and direct regression at different stages. Note that all of the recent neural-network-based facial landmark detection algorithms clearly show, that information explicitly present in the dataset is insufficient. To solve this problem several approaches are proposed:
\begin{itemize}
  \item use of an auxiliary representation, which contains structural information about the face, such as: 3D face mesh (DeFA); deformable shape model (LDDMM-Face); graph-based message passing (LAB); yaw, pitch, roll rotation angles (PFLD); landmark visibility (LUVLi); face representation as a graph model (DAG); offsets to anchors defined for faces with different poses (AnchorFace) or offset regression from neighboring landmarks (PIP);
  \item boundary representation either explicitly (LAB) or via attention module (PropagationNet, ADNet);
  \item hard example mining during training. Different variations on the theme have been presented in Wing, PFLD and PropagationNet papers;
  \item aggregating predictions for multiple input images: with style modification (SAN); after adversarial attack (GEAN); or with several different augmentations (SubpixelHeatmap). As it has been noted, minor changes in image might result in major shift in landmark prediction, averaging such predictions results in improved accuracy;
  \item train set augmentation using style (AVS) or adversarial attacks (SAAT);
  \item reduction of very large errors (outliers) and increased contribution of small to medium-sized errors (to better refine predictions): Wing, AWing and derivate works;
  \item subpixel heatmap precision (reducing heatmap quantization error): weighted argmax (LUVLI), joint heatmap and direct regression (PIPNet), global soft-argmax (ADNet), CNN- or Transformer-based refinement (HIH), local soft-argmax (SubpixelHeatmap).
\end{itemize}

\begin{table*}[tp]
  \caption{Face landmark detection method brief summary. The following abbreviations are used for algorithm type: D for direct regression, H for heatmap, SM for shape model, H~+~D indicates combined heatmap and direct methods.}\label{tab:landmark-detection-method-summary}
  \centering
  \begin{tabular}{lcp{4.2cm}p{6.8cm}}
    \toprule
    Model & Type                                                                                                                                                                               & Main Contribution & Notes \\
    \toprule
    ERT
          & D
          & First use of Ensemble of Regression Trees
          & Fast on CPU. Mediocre quality
    \\
    \hline
    DeFA
          & SM
          & 3D face mesh for faces with large pose and occlusion
          & Can train on datasets with different annotation schemes
    \\
    \hline
    SAN
          & H
          & Style neutralization
          & Good for landmark prediction under extreme lighting. Has slow inference
    \\
    \hline
    LAB
          & H\,+\,D
          & Boundary intermediate representation
          & Boundary module can be trained on datasets with different annotation schemes
    \\
    \hline
    AVS
          & H
          & Dataset augmentation via styled image generation
          & Idea is applicable to many methods
    \\
    \hline
    Wing
          & D
          & Special loss for direct regression
          & Training reduces small-to-medium errors. Can be adapted to any direct regression model
    \\
    \hline
    PFLD
          & D
          & Novel train loss and face angle prediction scheme
          & Good speed/quality ratio. The only method tested on a mobile device
    \\
    \hline
    FAN
          & H
          & Binarized convolutions for landmark prediction
          & Lacks testing on widespread face landmark datasets. Used in derivative works
    \\
    \hline
    AWing
          & H
          & Special loss for heatmap regression
          & Produces sharper heatmaps
    \\
    \hline
    MobileFAN
          & H
          & Network distillation
          & One of the fastest heatmap regression methods. Not tested on mobile devices
    \\
    \hline
    GEAN
          & H
          & Train/test-time augmentation using adversarial attack on face landmarks
          & Consolidates knowledge from face recognition model and landmark detection datasets. Requires multiple passes over the main network
    \\
    \hline
    HRNetV2
          & H
          & Improvement of HRNet architecture
          & HRNet reduces required computation over standard Hourglass
    \\
    \hline
    LUVLi
          & H
          & Prediction uncertainty estimation
          & Predicts landmark location and confidence jointly. Learning prediction confidence requires special dataset annotation. Application of CU-Net backbone for face landmark prediction
    \\
    \hline
    DAG
          & D
          & Topology-Adapting Graph Convolutional Network cascade
          & Captures face structural information. Good prediction for complicated pictures
    \\
    \hline
    PropagationNet
          & H
          & Boundary attention module. Focal Wing Loss
          & SOTA on 300W
    \\
    \hline
    SAAT
          & H
          & GAN and adversarial-attack-based training image generation
          & The idea can be applied to any method
    \\
    \hline
    LDDMM-Face
          & SM
          & LDDMM shape model with deep neural networks
          & Good dense landmark prediction after training on sparse landmark annotation
    \\
    \hline
    AnchorFace
          & D
          & 2-step prediction: anchor estimation; refinement with regression offsets and confidence scores
          & While the backbone is lightweight, inference time is still high
    \\
    \hline
    PIPNet
          & H\,+\,D
          & Coarse heatmap refined via direct regression
          & In theory, faster inference time with good quality. In practice, time is still high
    \\
    \hline
    ADNet
          & H
          & Point-edge heatmaps. Separate tangent/normal errors
          & SOTA on multiple datasets
    \\
    \hline
    HIH
          & H
          & Reducing heatmap quantization error via nested heatmaps
          & Simpler implementations (like SubpixelHeatmap) seem to work better
    \\
    \hline
    SubpixelHeatmap
          & H
          & Reducing heatmap quantization error via local soft-argmax
          & SOTA results on many benchmarks. Requires multiple passes over the network
  \end{tabular}
\end{table*}

\begin{figure}[htb]
  \centering
  \includegraphics[width=\linewidth]{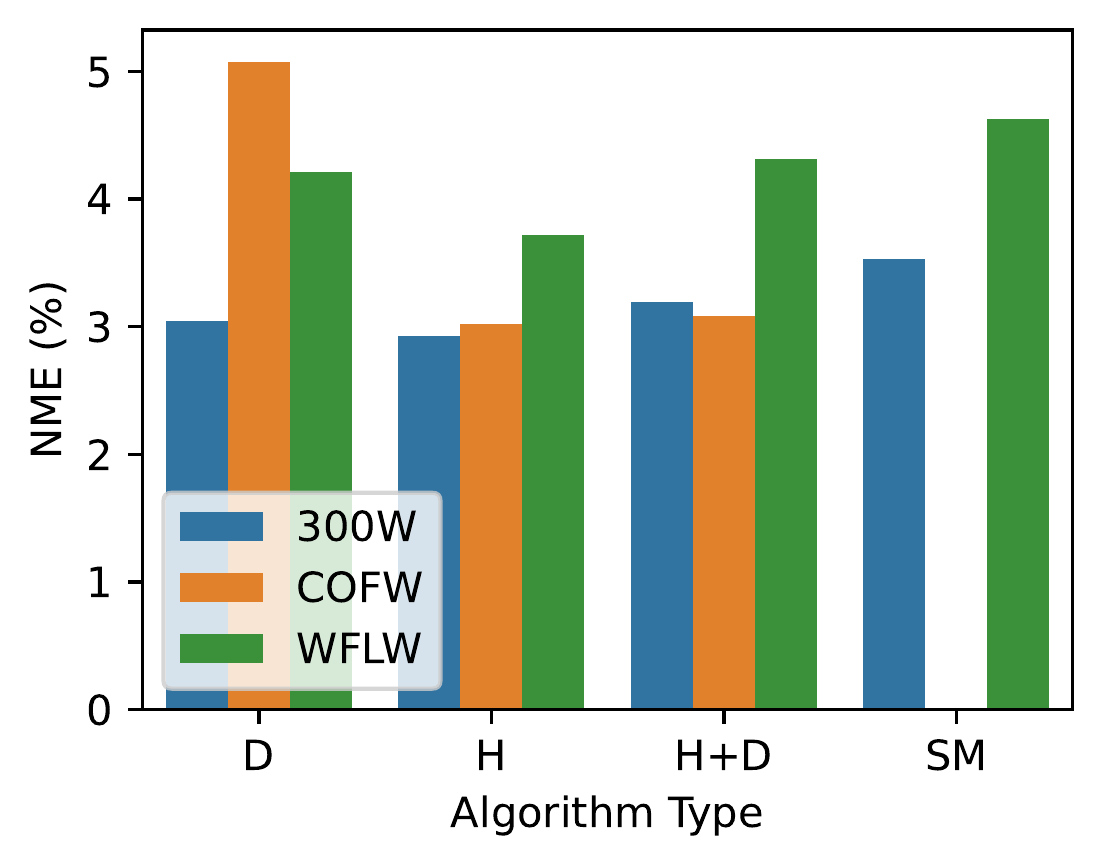}
  \caption{The best normalized mean error for each algorithm type: direct (D), heatmap-based (H), combined heatmap and direct (H+D), shape-model-based (SM). Heatmap-based approaches offer the best quality. Note, similar performance of direct and heatmap-based approaches on 300W.}\label{fig:nme-by-algorithm-type}
\end{figure}

\begin{figure}[htb]
  \centering
  \includegraphics[width=\linewidth]{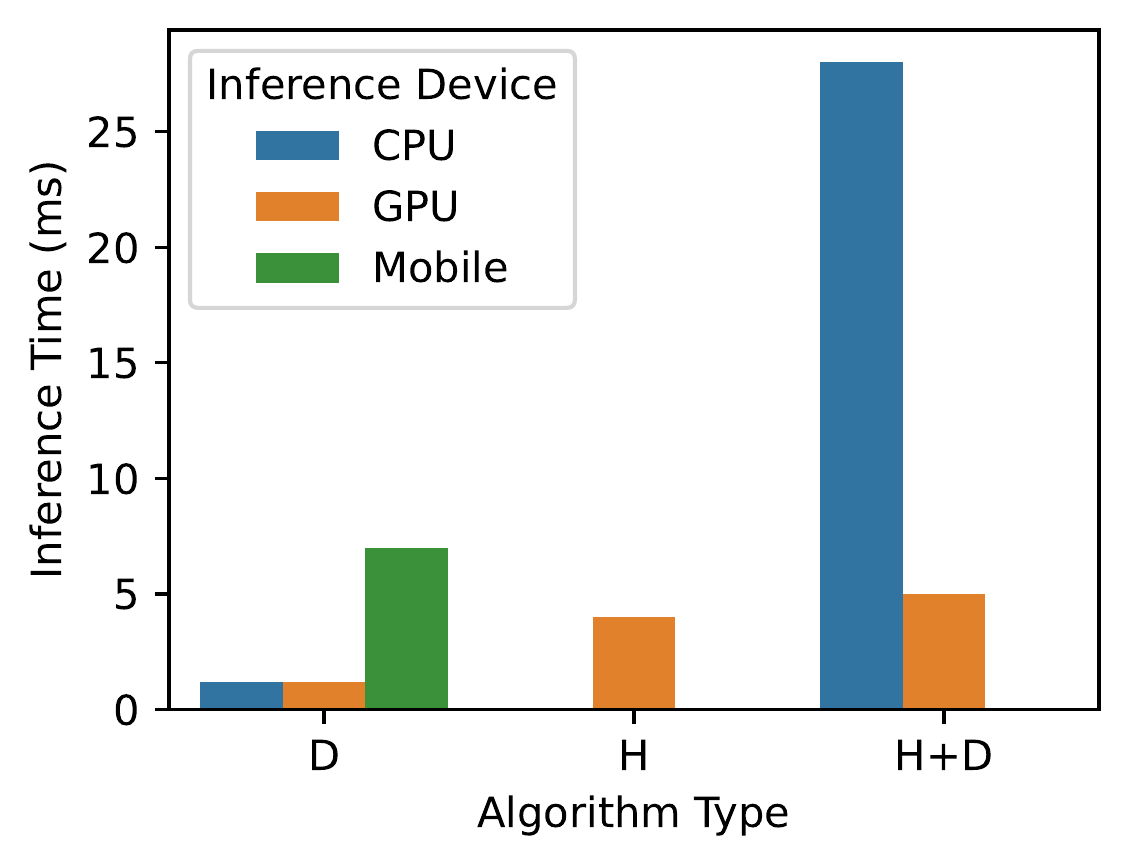}
  \caption{The best inference time for each algorithm type: direct (D), heatmap-based (H), combined heatmap and direct (H+D). Results are shown for different inference devices: CPU, GPU and Mobile. Direct regression algorithms have the best speed. Mixed H+D algorithm time is unexpectedly high.}\label{fig:inference-time-by-algorithm-type}
\end{figure}

\begin{figure*}[t]
  \centering
  \includegraphics[width=\linewidth]{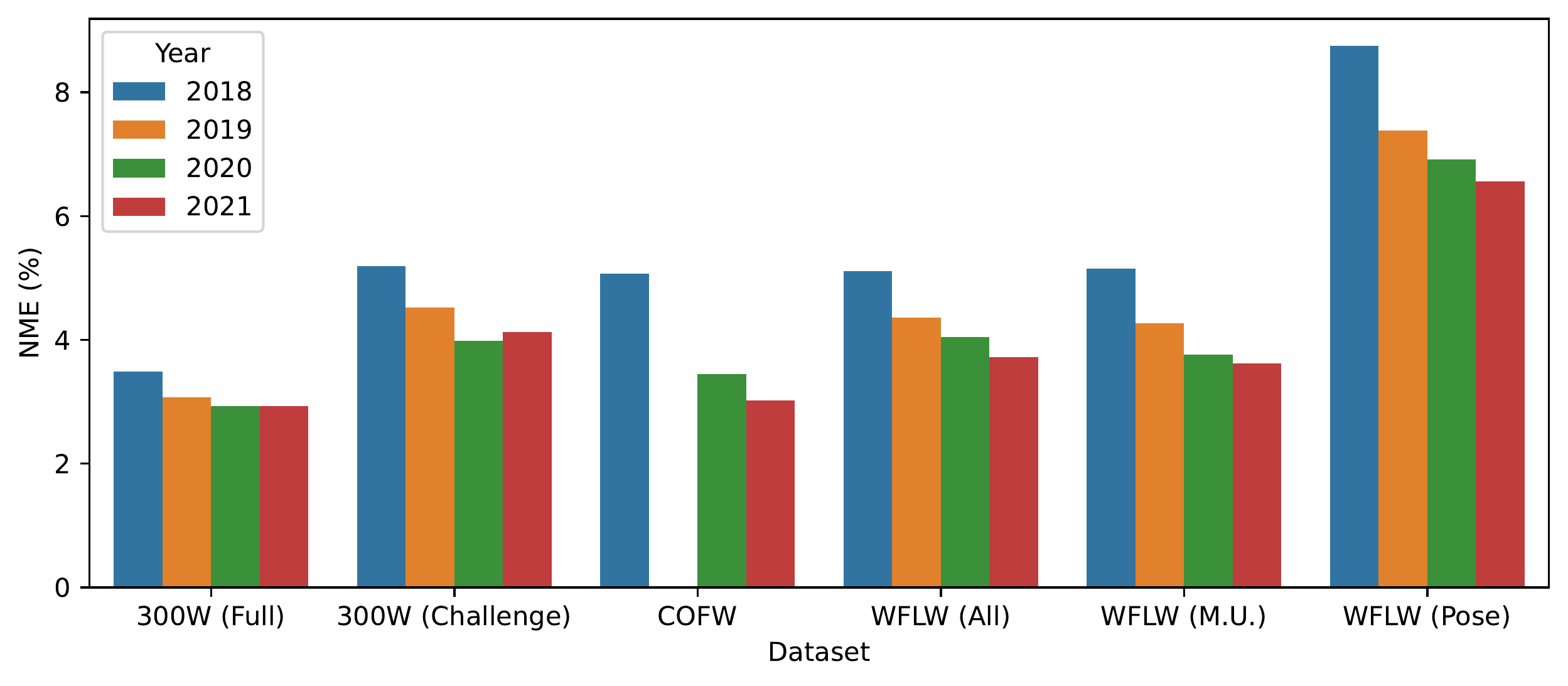}
  \caption{State-of-the-art normalized mean error (NME, \%) on 300W (Full, Challenge), COFW, WFLW (All, Make-Up, Pose) by years. Score on 300W dataset doesn't improve in 2021. Error on WFLW dataset, in contrast, significantly decreases with the time. WFLW error is higher (especially on images with large pose), than that of 300W and COFW, which makes it the most challenging dataset.}\label{fig:dataset-nme-by-year}
\end{figure*}

\begin{figure}[htb]
  \centering
  \includegraphics[width=\linewidth]{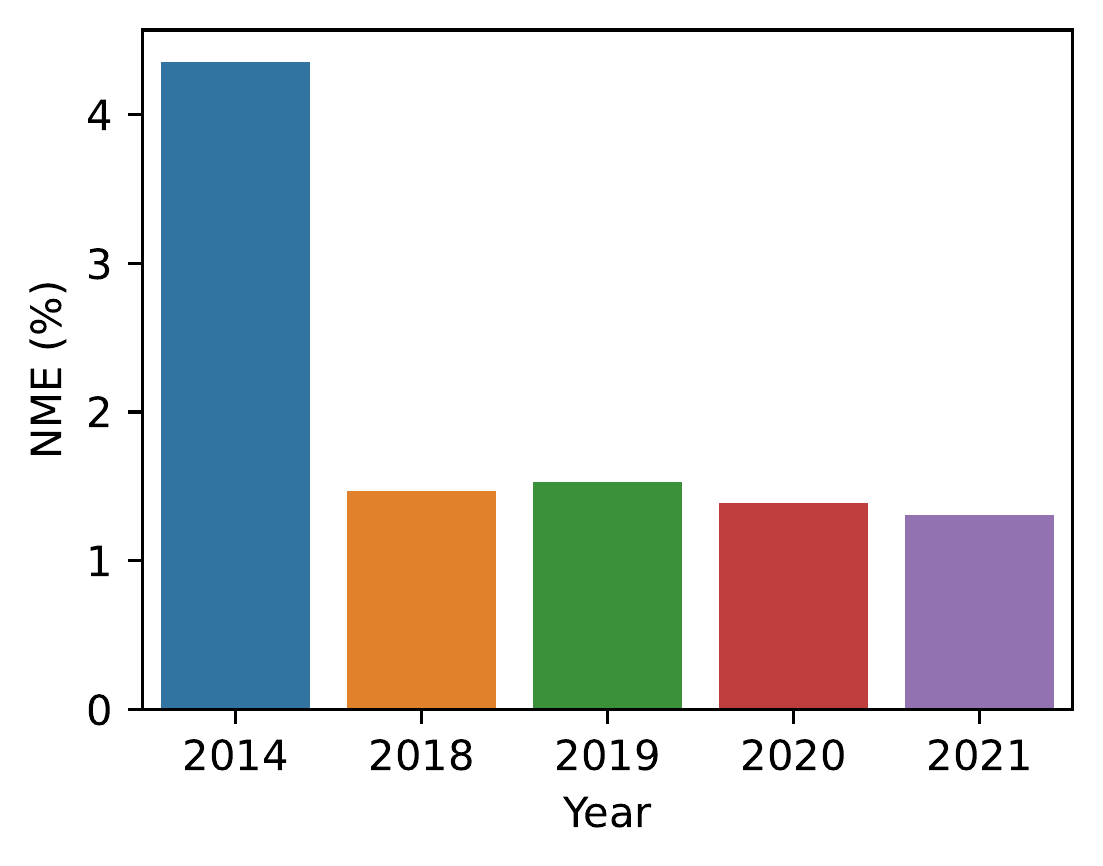}
  \caption{Algorithm Normalized Mean Error (NME, \%) on AFLW dataset by year. We deliberately left NME from 2014 to illustrate significant progress over nearly a decade. Note that NME still decreases year over year.}\label{fig:dataset-aflw-nme-by-year}
\end{figure}

In \cref{fig:nme-by-algorithm-type,fig:inference-time-by-algorithm-type,fig:dataset-nme-by-year,fig:dataset-aflw-nme-by-year} we present visual summary of the above-described tables, and discuss algorithms and datasets. Note, that for consistent comparison between datasets we have selected results with inter-ocular normalization, such results are present for 300W, COFW and WFLW datasets. Unfortunately, AFLW protocol specifies only face box diagonal normalization, which makes impossible direct comparison with other datasets. Because-of that we compare it separately. As usual, we do not include results with extra training data used.

In \cref{fig:nme-by-algorithm-type} we show the best normalized mean error achieved by each type of algorithms: direct, heatmap-based, combined heatmap and direct, shape-model-based. All of these algorithms are based on neural networks. Direct and heatmap-based approaches have nearly the same performance on 300W (Full) dataset, with a slight advantage of heatmap-based approaches. In contrast, heatmap-based approaches show WFLW performance significantly better, than direct regression algorithms. The best direct NME of 4.21 \% is achieved by DAG algorithm (direct prediction based on graphs), the best heatmap-based NME is much lower at 3.72 \%. This is also state-of-the-art result, that is achieved by SubpixelHeatmap. Note, significantly higher COFW error for direct approaches. This happens because only a single direct regression algorithm has been tested on COFW, that is Wing algorithm. The approach is quite old. Hence, this error spike should be considered as an outlier. Next, we take a look at combined heatmap-direct approaches. There are only two of them, LAB (which is quite old) and PIPNet. PIPNet proposed to predict coarse heatmap and then refine it via direct regression. While the idea is quite promising, the algorithm offers worse performance than direct or heatmap-based approaches. Finally, neural-network-based shape model algorithms offer the worst performance. Such approaches are quite useful, when network training is performed on images with different annotation schemes, or when 3D face mesh is required. Note, in this survey only DeFa algorithm produces 3D mesh. However, for most other use cases such approaches should not be considered.

In \cref{fig:inference-time-by-algorithm-type} we show the best inference time achieved by algorithms of each type. We do not visualize results for shape-model-based approaches as no timings have been presented in the corresponding papers. As expected, the fastest approaches use direct regression. They typically have more lightweight backbones. Also, they do not need to predict large heatmaps for each of the landmarks, which saves computation. Heatmap-based approaches have the best GPU inference time at 4.0 ms (achieved by MobileFan (0.5)), which is worse than 1.2 ms achieved by direct approach (PFLD 0.25X). While the timings seem to be quite small, note that both approaches offer significantly lower accuracy, than state-of-the-art algorithms. State-of-the-art algorithms still execute around 100 ms on GPU. Only GPU timings are available for heatmap-based approaches. Finally, we show timings for combined heatmap and direct regression methods. The best results achieved by PIPNet with ResNet-18 backbone. While the backbone is quite lightweight, inference time of 28.0 ms on CPU and 5.0~ms on GPU for this model is quite high. We would have expected the timings to be lower here.

In \cref{fig:dataset-nme-by-year} we show the best algorithm performance grouped by dataset and year. To begin with, we discuss 300W dataset results, which are presented for Full and Challenge sets. We note that no progress has been made in 2021 in comparison to 2020. We expect error of all algorithms to stop decreasing at some point, as both training and test sets contain annotation errors. Thus, it would be interesting to see, whether any progress is made in 2022, or the remaining algorithm error is due to incorrect annotation. On COFW significant progress has been made over years. And WFLW dataset is still the most challenging. WFLW dataset has been introduced in 2018, and the largest improvement has been achieved in the following year, which is especially obvious on Pose subset, that has the highest normalized mean error. As discussed before, annotation of faces with unusual Make-Up has the least challenge to facial landmark algorithms, but NME has decreased even on this subset. Moving over to AFLW, in \cref{fig:dataset-aflw-nme-by-year} we show AFLW performance over years. The performance is still being improved. We have deliberately left NME for algorithm from 2014 to note the significant progress made in 7 years.

To conclude this section, we would like to note the best algorithms on different datasets: SubpixelHeatmap, ADNet, PropagationNet. All of these algorithms are heatmap-based. The key ideas proposed in them are complimentary, for instance, SubpixelHeatmap has offered a way to decrease heatmap quantization error. Recall, that input image and landmark annotations are of resolution \(256 \times 256\), while the heatmap is only \(64 \times 64\). The authors of ADNet have presented Point-Edge heatmaps as attention masks, and PropagationNet has presented Focal Wing Loss modification. However, inference time of ADNet is still quite high at 95.29 ms on GPU. We expect PropagationNet to be even slower, based on presented in \cref{tab:backbone-inference-speed} number of floating point operations. SubpixelHeatmap neural network is inferred for each image several times, which will also be slow. Thus, we would like to see faster algorithms in future, and those that are easily applicable to mobile devices.

\section{Facial Landmark Detection: Applications}\label{sec:landmark-detection-applications}

\subsection{Mobile-Friendly Joint Face and Landmark Detection}\label{sec:joint-face-landmark-detection}

As we have noted previously, face landmark detection is one face processing pipeline steps. To actually get a dense landmark annotation, face has to be detected first and cropped based on its bounding box. In this section we present some of mobile-friendly face detection methods. Interestingly, these methods also predict coarse (5 or 6) face landmarks, such as eyes, mouth, nose.

\textbf{Multi-task Cascaded Convolutional Networks (MTCNN)}~\cite{MTCNN}. The neural network is trained jointly to detect faces and landmark locations (five of them, to be precise: eyes, tip of the nose, mouth corners), which improves quality on both tasks. The network is built in a form of a three-network cascade: Proposal Network (P-Net), Refine Network (R-Net), Output Network (O-Net). Each network predicts face bounding rectangle, probability that a particular rectangle contains a face, and five landmarks. P-Net is a fast fully convolutional network, which processes the original image in multiple resolutions (the so-called image pyramid). This network outputs a lot of coarse face rectangle predictions, which are then filtered out by the Non-Maximum Suppression (NMS) algorithm. Subsequently, R-Net refines the predicted rectangles. It does so without reprocessing the whole image, which saves computation time. NMS is then applied again. Last, O-Net makes the final refinement. This is the slowest network in the cascade, but it processes a small number of face rectangles. According to the authors, to improve quality it is important to solve the following tasks at the same time: 1) classify bounding rectangle as a face or not a face; 2) perform regression over bounding rectangle coordinates; 3) localize face landmarks. Each task has a weight $\alpha$ assigned: for P-Net and R-Net $\alpha_1 = 1, \alpha_2 = \alpha_3 = 0.5$, for O-Net $\alpha_1 = 1, \alpha_2 = 0.5, \alpha_3 = 1$ correspondingly. At training time online hard-example mining has been used, meaning that training is performed on complicated examples while skipping those, on which network prediction is quite accurate already. In the paper the authors select around 70\% of hardest examples in each training batch.

\textbf{BlazeFace}~\cite{BlazeFace} is a novel approach to joint face and landmark detection. 6 landmarks are predicted: eye center, ear tragions, mouth center, and nose tip). The algorithm was specifically designed for inference on mobile devices. The authors claim sub-second detection time on mobile for Tensorflow~\cite{Tensorflow} GPU implementation. The approach is based on Single Shot Detector (SSD)~\cite{SSD} with MobileNetV2 backbone. The authors propose to modify MobileNetV2 to improve performance to accuracy ratio. For that they increase complexity of Bottleneck block (main building block of MobileNetV2 architecture) and decrease the number of such blocks at the same time. Also, they have optimized SSD architecture for face detection by removing the ability to predict wide or tall bounding boxes, that are not common for faces. In addition, intra-frame jitter produced by the NMS algorithm has been reduced via a separate bounding box regression module. The network is proposed in 2 configurations: one for pictures taken on back camera (typically smaller faces) and another for frontal camera photos (typically larger faces). While the network has good accuracy, the input resolutions are fixed to \(128\times128\) or \(256\times256\), which is a disadvantage of the method. MTCNN, for example, can take images of arbitrary resolution as input. Note, that training has been performed on a closed dataset. Thus, it is not possible to reproduce the results.

A comprehensive review of other modern face detection methods is presented in~\cite{FaceDetectionSurvey}. We see the following prospects for further research in the field of joint face and landmark detection:
\begin{itemize}
  \item inference speed to accuracy ratio requires improvement. Faster approaches often have lower quality;
  \item large annotated dataset is required to train the model. If the dataset is biased (unbalanced) in race or gender, face detection accuracy of underrepresented groups will typically suffer.
\end{itemize}

\subsection{Face Animation and Reenactment}\label{sec:face-animation}

Facial landmark detection is used in human or imaginary character face animation algorithms. Applications include actor animation in movies, creation of TV or game virtual newscasters (as a 3D model or directly via GAN image generation). Recent landmark detection algorithms enable this without costly equipment by using a simple RGB camera.

According to research presented in a series of papers, movie dubbing process from foreign languages is expensive and time-consuming. This is because lip movement for the original and dubbed audio tracks should match. Furthermore, the movement discrepancy leads to discomfort when watching movies, especially for hearing-impaired people. As a solution, authors of~\cite{VDub} propose to change lip movement during the dubbing process. Their algorithm detects facial landmarks and substitutes mouth region with a 3D model, adapted for the speaker. However, at this stage the substitution is still visible. Besides that, DeFA algorithm can build a 3D whole-face mesh for varied poses and emotions, as has been said previously.

\begin{figure}[htb]
  \captionsetup[subfigure]{justification=centering}
  \centering
  \begin{subfigure}{0.24\linewidth}
    \centering
    \includegraphics[height=0.9\linewidth]{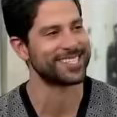}
  \end{subfigure}
  \begin{subfigure}{0.24\linewidth}
    \centering
    \includegraphics[width=0.9\linewidth]{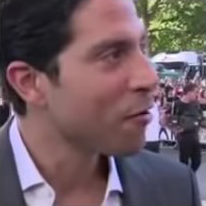}
  \end{subfigure}
  \begin{subfigure}{0.24\linewidth}
    \centering
    \includegraphics[width=0.9\linewidth]{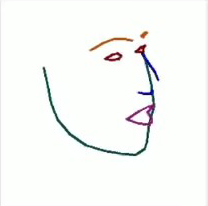}
  \end{subfigure}
  \begin{subfigure}{0.24\linewidth}
    \centering
    \includegraphics[width=0.9\linewidth]{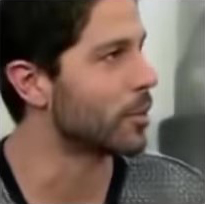}
  \end{subfigure}
  \caption{Reenactment scheme. (a)~source character image (the one we want to reenact); (b)~one of frames of driver actor; (c)~extracted facial landmarks that are fed to the reenactment algorithm; (d)~reenactment result.\protect\footnotemark}\label{fig:talking-heads}
\end{figure}

Many of the recent neural-network-based algorithms do not use an intermediate 3D face model for realistic image generation, but generate images directly from facial landmark locations via Generative Adversarial Network (GAN). For instance, the authors of~\cite{ReenactmentRu} by using MAML~\cite{MAML} meta-learning approach, GAN and the so-called perceptual loss~\cite{PerceptualLosses}, obtain high face reenactment quality (\cref{fig:talking-heads}). Landmark information extracted from an image is one of the neural network inputs. FAN algorithm is used to extract the landmarks. The algorithm has some disadvantages though. For instance, when actor, that drives the animation, has significantly different face shape from animated face, the resulting animation is unrealistic and contains artifacts. According to the authors' report, this method outperforms the competition for face emotion transfer task in few- or one-shot problem statement. The authors note, that an improvement of facial extraction algorithm and addition of gaze direction might have improved the reenactment quality.

\footnotetext{Distributed under Creative Commons -- Attribution 3.0 license. Based on \href{https://youtu.be/p1b5aiTrGzY}{this source}.}

In~\cite{NeuralLipSync} authors are using Pix2PixHD~\cite{Pix2PixHD} neural network to accomplish lip sync task. It has been proposed to synthesize the intermediate face representation using its boundaries, face landmarks (using Dlib library) and sound-track-based representation.

In another work FReeNet~\cite{FReeNet} algorithm is presented for reenactment between different (unknown during training) people. For that a special Unified Landmark Converter module has been introduced, which adapts facial landmark coordinates between different people. Landmarks for the source and target people are extracted via PFLD algorithm. Then images are generated via Cycle-GAN~\cite{CycleGAN} and a special loss function. The use of landmark converter module has given the largest performance increase on the test sets.

A survey of emotion transfer and face reenactment methods can be found in~\cite{DeepfakesSurvey} Section ``Expression Swap''. Most recent algorithms, that focus on face animation of real people use generative adversarial neural networks (GANs). Currently, these approaches have the following limitations, that require a solution in future, for instance:
\begin{itemize}
  \item videos produced by neural networks lack temporal stability. For instance, face animation might jitter, artifacts may appear on the screen;
  \item face animation under large pose might cause unrealistic face deformation;
  \item animation quality significantly suffers, if source character and animation driver have different face shape.
\end{itemize}

\subsection{Driver Status Tracking}\label{sec:driver-tracking}

A large number of car accidents happens because-of sleepy or tired drivers. Expensive cars offer capabilities of emergency stopping when an obstacle is detected, and line-keep assist. Mainstream cars do not have such features. In both cases, it is better to track driver status and stop the car early, than to apply emergency brakes. Most of the research in the field is focused on implementing status tracking in an autonomous way (without Internet connection). Driver's smartphone or low-power portative device (such as Raspberry Pi) is used to process video signal from a camera placed in a car's cabin. Neural-network-based algorithms are among the most widely used approaches here.

In~\cite{DriverDrowsinessDetection} the authors estimate driver tiredness via a neural network that takes facial landmarks as an input. Driver's face and landmarks are detected with existing methods. In contrast, in~\cite{MobileDriverDrowsiness3dNN} a MobileNetV2-based architecture is presented to estimate driver's sleepiness directly from the video stream (without an intermediate step of landmark detection), yet total training time is quite high. In~\cite{LightweightDriverMonitoringSystem} neural-network-based landmark detection is utilized to simplify dataset labelling, then a different network is trained to recognize driver's status. In addition to fatigue, the authors also estimate driver's distraction by tracking whether he looks in safe zones (such road, rear-view mirror, dashboard, etc.) or not. In~\cite{DriverStatusIsAutonomousDrivingEra} a system that tracks driver's ability to take over the driving from level 2 autonomous cars (partial driving automatization) is studied. The authors acquire driver's video via an infrared camera. Decision whether the driver is distracted is based on the detected landmarks. These and similar algorithms are developed to make the roads safer.

Special hardware can also be used to track driver status~\cite{DriverDrowsinessDetection,MobileDriverDrowsiness3dNN}, for instance, tracking of driving wheel movement; wearable devices that perform Electrocardiography (ECG) and heartbeat measurements. However, both of these approaches are more expensive and cannot track driver's distraction from the road.

Neural-network-based driver status tracking algorithms have the following limitations, for instance:
\begin{itemize}
  \item achieving sufficiently fast inference on a mobile device is a challenge;
  \item driver status tracking is often performed at night-time when driver is poorly illuminated. Facial landmark detection accuracy suffers in such conditions, especially given limited computation power.
\end{itemize}
This is why development of mobile networks and landmark detection algorithms will definitely enhance the quality of driver status tracking systems.

\subsection{Face Recognition and Emotion Classification}\label{sec:face-recognition}

To begin with, we briefly talk about algorithms that perform one of the following tasks (often, the same algorithm can perform all of them): 1) face verification, when 2 pictures are given and the task is to say whether they contain the same person; 2) face recognition: given a photo and a known person database, algorithm should say who is on the photo or that the person is unknown; 3) clusterization, where the task is to group similar faces. The most efficient algorithms use face preprocessing, that is face detection and tight crop. Often for improving recognition quality the so-called ``face alignment'' should additionally be performed, that is a geometrical image transformation, when facial landmarks are moved to the canonical locations. Many of the modern algorithms use MTCNN for joint face detection and localization of 5 landmarks. The topic of face recognition is well-described in, for example,~\cite{DeepFaceRecognitionSurvey}. We note in particular, high interest to face recognition directly on mobile devices~\cite{MobiFace, MobileFaceNets}.

Also, we discuss emotion recognition. Our emotions mostly consist of lip, eyes, eyebrows or mouth movements. That is why in certain cases it is fruitful not to force the neural network to learn face parts during emotion recognition on its own, but to feed this information detected by another algorithm together with the original image~\cite{FacialEmotionRecognitionSurvey, DeepFacialExpressionRecognitionSurvey}.

The field of face recognition has several problems that require a solution in future, for example:
\begin{itemize}
  \item face recognition when photos represent a person of different ages;
  \item when face occlusion is significant, which is especially important when medical masks have become common;
  \item faces with large pose and emotion;
  \item also, some of the backbones discussed here in a context of facial landmark detection are used for face recognition and emotion classification. Thus, improving neural network backbones is important as well.
\end{itemize}

\section{Facial Landmark Detection: Vulnerabilities}\label{sec:vulnerabilities}

Modern computer vision algorithms (including neural networks) are amenable to the so-called ``adversarial attacks'', first reported in the field of computer vision in ~\cite{IntriguingPropertiesOfNN}. The authors were able to drastically change neural network prediction in classification task by adding especially crafted noise (invisible to human eye) to an image. The attack has been conducted by maximizing network error on the target image via L-BFGS method. Images with adversarial noise are almost always misclassified on MNIST dataset. It should be stressed that during the adversarial attack the network itself is not modified, only the images fed to it. Moreover, adversarial examples often remain malicious to networks different from the one they were crafted for, given that another network was trained on the same or similar dataset. It should be noted, that adding random noise has much lower negative effect on the network's classification accuracy, than adversarial attack noise. In~\cite{ScopingAdversarialAttack} it has been shown that for a successful adversarial attack on the MNIST dataset, model as simple as logistic regression can be used to generate adversarial examples. The attack remains efficiently transferrable to architectures, that are more complicated.

If previous algorithms have attacked a digital image (stored in computer memory), in~\cite{AdversarialExamplesInPhysicalWorld} it has been shown that attacks can be performed through a smartphone camera. In~\cite{MinimizingPerceivedImageQualityLossArxiv} binary importance maps have been introduced, which hint where adversarial marks should be placed on a piece of paper to fool the network trained to classify handwritten digits. The first adversarial attacks were white-box, i.e., the network architecture and trained weights are known to the attacker. Follow-up works similar to~\cite{ZOO} and others have shown that it is possible to perform black-box attacks without such knowledge. Despite the fact that numerous works are devoted to detecting or preventing attacks from happening, new more advanced algorithms bypass all of the defense methods~\cite{BypassingTenDetectionMethodsCarliniWagner}. A survey of adversarial attack methods can be found in~\cite{AdversarialAttackSurvey}. All of them are applicable to algorithms of face or facial landmark detection.

In the meantime, there exist special methods that can prevent the face from being found or correctly detected by using stickers or accessories in real world. In~\cite{AccessorizeToCrime} it has been shown, that in a controllable environment it is possible to fool face recognition algorithm or Viola-Jones face detector. The authors used special eyeglasses with a print on a frame. In~\cite{MtcnnAdversarialAttack} it has been proposed to fool MTCNN face detection algorithm with the use of stickers on cheeks or medical mask. In cases when the face cannot be detected, landmark localization cannot be performed either. Face recognition adversarial attack based on facial landmarks is presented in~\cite{FacesAdversarialAttack}.

\section{Conclusion}

From a detailed survey, we see the following facial landmark detection algorithm problems, that require a solution in future research: 1) despite a significant growth of methods' quality, few of them focus on the real-world applicability in resource-constrained environments, such as mobile or edge devices; 2) many applications require high performance on mobile or portable devices, yet to the best of our knowledge, authors of only a single algorithm have targeted a mobile application directly in the original paper. Note that state-of-the-art algorithms have slow inference speed; 3) while modern research already focuses on datasets in uncontrollable environments, a promising research direction is to enhance algorithms in even harsher conditions, for images with large pose and significant face occlusion, while still maintaining high landmark density. Error of current generation of algorithms in these conditions is quite high. We see WFLW dataset as the one posing the most interest for further research. Also, it would be desirable to see more of the novel facial landmark detection algorithms to report their inference speed on desktop GPU, and if possible, on mobile devices.

We hope, that the described modern developments in all of the sections will lead the reader to new ideas of practical use and further research directions.

\begin{small}

  \subsubsection*{Competing interests}

  The authors have declared that no competing interests exist.

  \subsection*{Funding}

  The work is supported by the state budget scientific research project of Dnipro University of Technology ``Development of New Mobile Information Technologies for Person Identification and Object Classification in the Surrounding Environment'' (state registration number 0121U109787).

  \subsection*{Authors' contributions}

  KK proposed the idea, selected impactful literature in the field, surveyed and analyzed the literature, wrote the manuscript; LK validated the manuscript. All authors read and approved the final manuscript.

  \printbibliography

\end{small}

\cornersize{.2}
\setlength{\fboxsep}{8pt}

\ovalbox{

  \begin{minipage}{7cm}
    \textbf{Citation:} \nohyphens{K.~Khabarlak, L.~Koriashkina \emph{Fast Facial Landmark Detection and Applications: A Survey}. Journal of Computer Science \& Technology, vol. 22, no. 1, pp. 12–41, 2022.}

    \textbf{DOI:} \href{https://doi.org/10.24215/16666038.22.e02}{10.24215/16666038.22.e02}
  \end{minipage}
}

\end{document}